\documentclass[10pt,twocolumn,letterpaper]{article}

\usepackage[draft,layout={footnote,noinline}]{fixme}
\usepackage{cvpr}
\usepackage{times}
\usepackage{epsfig}
\usepackage{graphicx}
\usepackage{amsmath}
\usepackage{amssymb}
\usepackage{stfloats}
\usepackage{subcaption}
\usepackage{url}
\usepackage[utf8]{inputenc}
\usepackage[table,xcdraw]{xcolor}
\usepackage{multirow}
\usepackage[normalem]{ulem}

\usepackage{algorithm}
\usepackage{algorithmic}
\DeclareMathOperator*{\argmax}{argmax}

\newcommand{\sectionref}[1]{Section~\ref{sec:#1}}
\newcommand{\figref}[1]{Figure~\ref{fig:#1}}
\newcommand{\tableref}[1]{Table~\ref{tab:#1}}
\newcommand{\equationref}[1]{Equation~(\ref{eq:#1})}
\newcommand{\STAB}[1]{\begin{tabular}{@{}c@{}}#1\end{tabular}}

\usepackage{titling}

\cvprfinalcopy 

\def\cvprPaperID{9675} 

\ifcvprfinal\fi
\begin{document}

\title{3D-ZeF: A 3D Zebrafish Tracking Benchmark Dataset}
\date{\vspace{-2ex}}
\author{Malte Pedersen\thanks{Equal contribution}~, Joakim Bruslund Haurum\footnotemark[1]~, Stefan Hein Bengtson, Thomas B. Moeslund\\
Visual Analysis of People (VAP) Laboratory, Aalborg University, Denmark\\
{\tt\small mape@create.aau.dk, joha@create.aau.dk, shbe@create.aau.dk, tbm@create.aau.dk}
}

\maketitle
\thispagestyle{empty}

\begin{abstract}
In this work we present a novel publicly available stereo based 3D RGB dataset for multi-object zebrafish tracking, called 3D-ZeF.
Zebrafish is an increasingly popular model organism used for studying neurological disorders, drug addiction, and more.
Behavioral analysis is often a critical part of such research.
However, visual similarity, occlusion, and erratic movement of the zebrafish makes robust 3D tracking a challenging and unsolved problem.

The proposed dataset consists of eight sequences with a duration between 15-120 seconds and 1-10 free moving zebrafish.
The videos have been annotated with a total of 86,400 points and bounding boxes. 
Furthermore, we present a complexity score and a novel open-source modular baseline system for 3D tracking of zebrafish.
The performance of the system is measured with respect to two detectors: a naive approach and a Faster R-CNN based fish head detector.
The system reaches a MOTA of up to 77.6\%. 
Links to the code and dataset is available at the project page \url{https://vap.aau.dk/3d-zef}
 
\end{abstract}

\section{Introduction}

Over the past decades, the use of zebrafish (\textit{Danio rerio}) as an animal model has increased significantly due to its applicability within large-scale genetic screening \cite{eisen_zebrafish_1996,haffter_identification_1996}.
The zebrafish has been used as a model for studying human neurological disorders, drug addiction, social anxiety disorders, and more \cite{collier_zebrafish_2014, lin_zebrafish_2016, soares_using_2018, kalueff_zebrafish_2014,meshalkina_zebrafish_2018, khan_zebrafish_2017}.
Locomotion and behavioral analysis are often critical parts of neuroscientific and biological research, which have traditionally been conducted manually \cite{li_dominant_1997,muller_swimming_2004,mcelligott_prey_2005}.
However, manual inspection is subjective and limited to small-scale experiments.
Therefore, tracking systems are getting increasingly popular due to their efficiency and objectivity.
The majority of the solutions has been developed for terrestrial animals or fish in shallow water, and most studies have been based on 2D observations in scientific \cite{fontaine_automated_2008,risse_fimtrack:_2017, ohayon_automated_2013, qian_automatically_2014, feijo_algorithm_2018,franco-restrepo_review_2019, romero-ferrero_idtracker.ai:_2019} and commercial systems \cite{noldus_ethovision:_2001, systems_lolitrack_2018, viewpoint_zebralab_nodate, systems_videomot2_2014}.
However, observations in a single plane cannot capture all the relevant phenotypes of fish \cite{kalueff_towards_2013, cachat_video-aided_2011, macri_three-dimensional_2017}.
\begin{figure}[!t]
    \centering
    \includegraphics[width=0.45\textwidth]{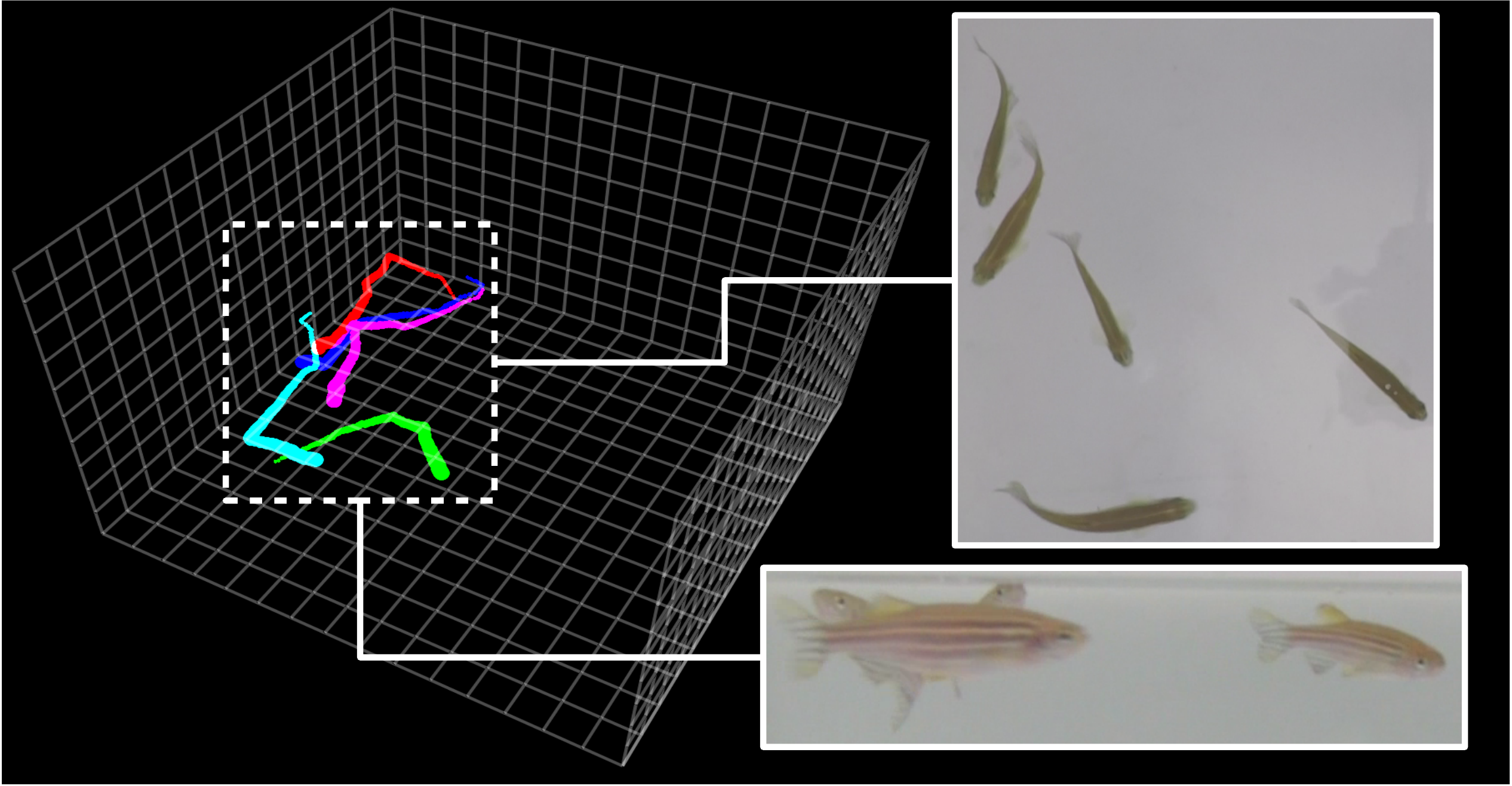}
    \caption{An example that illustrates the difference between the two perspectives. The 3D trajectories are estimated based on the head point annotations.}
    \label{fig:2fish_example}
\end{figure}
Estimating the 3D trajectories of multiple zebrafish accurately is difficult due to their erratic movement, visual similarity, and social behavior \cite{miller_quantification_2007}, see \figref{2fish_example}.
This may be one of the reasons why no commercial solution has been developed yet.
Only few groups in the scientific community have addressed the problem, focusing mainly on stereo vision \cite{alzubi_real-time_2015,cheng_obtaining_2018,qian_skeleton-based_2017, wang_3d_2016-1, qian_feature_2017} and monocular stereo using mirrors \cite{audira_simple_2018,xiao_research_2016}.
However, no labeled datasets have been made publicly available within the field, which makes a fair comparison between the applied methods difficult.
This ultimately hinders significant developments in the field as we have seen in other computer vision fields with common datasets.
Therefore, our contributions are
\begin{itemize}
    \item a publicly available RGB 3D video dataset of zebrafish with 86,400 bounding box and point annotations.
    \item an open-source modular baseline system.
\end{itemize}

A large part of 3D multi-object tracking methods are developed for LiDAR-based traffic datasets \cite{Menze2015ISA, Voigtlaender2019CVPR, caesar2019nuscenes, waymo_open_dataset, chang2019argoverse} or RGB-D tracking \cite{song2013tracking, lukezic2019cdtb}.
However, to the best of our knowledge, there exists no publicly available annotated RGB stereo dataset with erratic moving and similarly looking subjects like the one we propose.

\section{Related Work}\label{sec:related_work}
\begin{figure*}[t!]
    \centering
    \begin{subfigure}[b]{0.6\textwidth}
        \centering
        \includegraphics[width=\textwidth]{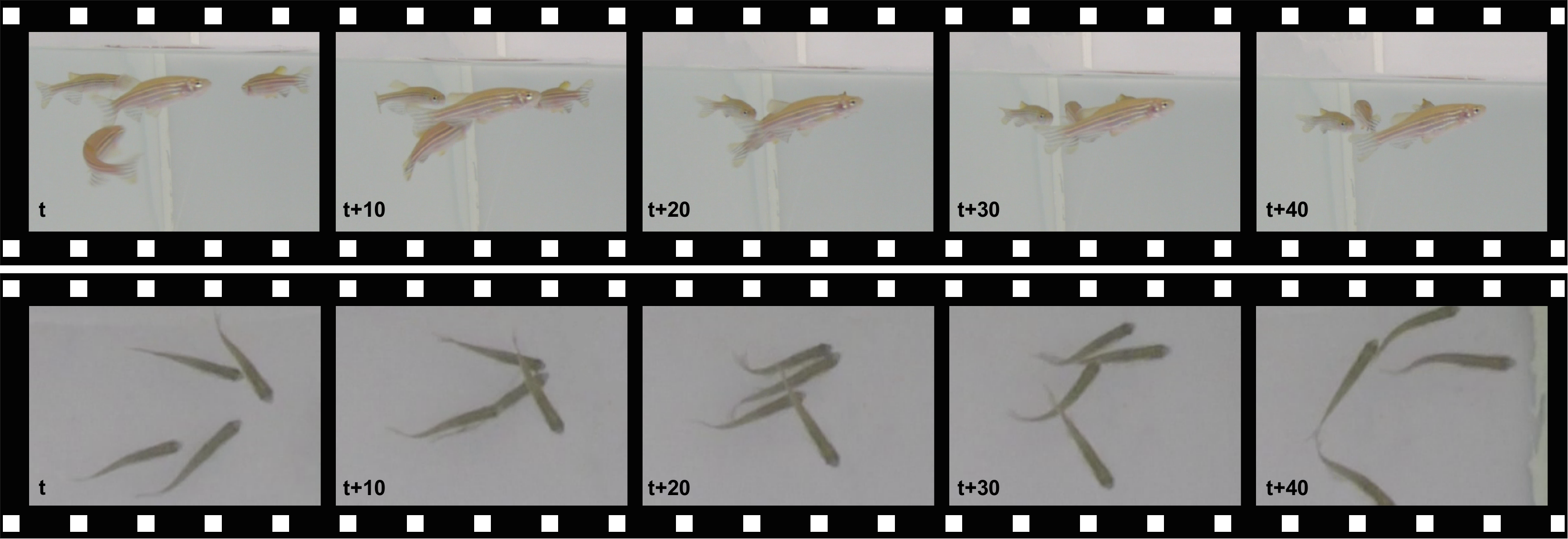}
    \end{subfigure}
    \quad\quad
    \begin{subfigure}[b]{0.3\textwidth}
        \centering
        \includegraphics[width=\textwidth]{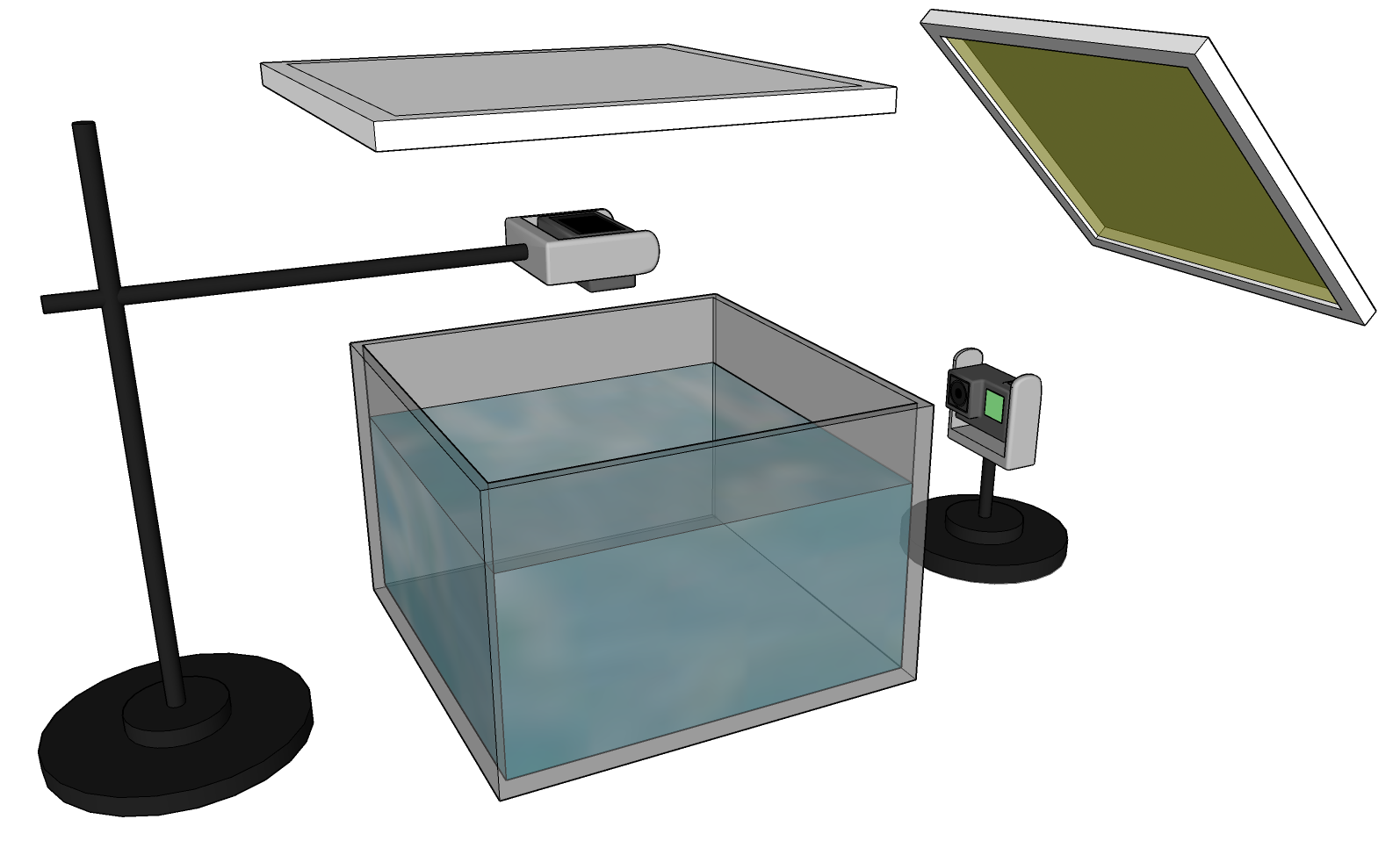}
    \end{subfigure}
    \caption{Five frames from two different occlusion scenarios. The upper frames are from the front-view and the lower frames are from the top-view. An illustration of the experimental setup is shown to the right.}
    \label{fig:examples}
\end{figure*}

\textbf{Multi-Object Tracking (MOT).} 
Reliably tracking multiple objects is widely regarded as incredibly difficult. 
The interest in solving MOT has been steadily increasing since 2015 with the release of the MOT \cite{MOTChallenge2015, MOT16, MOT19_CVPR}, UA-DETRAC \cite{UA-DETRAC_2017, UA-DETRAC_2020}, and KITTI \cite{Menze2015ISA, Voigtlaender2019CVPR} challenges. 
Within the MOT challenges, the current focus is on either aiming to solve the association problem using deep learning \cite{DAN_2019}, using techniques such as intersection-over-union based tracking \cite{IOUTrack_2018}, or disregarding tracking-specific models and utilizing the improvements within object detections \cite{Bergmann_2019_ICCV}.

\textbf{Zebrafish Tracking.} Vision-based tracking systems developed for studying animal behavior have traditionally been based on 2D \cite{romero-ferrero_idtracker.ai:_2019, perez-escudero_idtracker:_2014, sridhar_tracktor:_2019,monck_biotracker:_2018,rodriguez_toxtrac_2017,harmer_pathtrackr_2019, liu_tracking_2019} due to simplicity and because the movement of most terrestrial animals can be approximated to a single plane.
The majority of research in zebrafish tracking has followed this path by only allowing the fish to move in shallow water and assuming that motion happens in a 2D plane.

A 2D animal tracker, called idTracker presented by Perez-Escudero \etal in 2014 \cite{perez-escudero_idtracker:_2014}, uses thresholding to segment blobs and is able to distinguish between individual zebrafish based on intensity and contrast maps.
In 2019, Romero-Ferrero \etal presented an updated version of idTracker, called idtracker.ai \cite{romero-ferrero_idtracker.ai:_2019}, which is the current state-of-the-art 2D tracker system based on convolutional neural networks (CNN) for handling occlusions and identifying individuals.
The subjects are observed with a camera positioned above a tank with a water depth of 2.5 cm and the distance between camera and subjects is, therefore, approximately the same at all times.
As stated by the authors, this simplifies the task compared to a real 3D tracking scenario.

However, as most aquatic species move in three dimensions, trajectories in 3D are required to thoroughly describe their behavior \cite{zhu_catadioptric_2007,cachat_three-dimensional_2011}.
The most frequently used acquisition method when dealing with studies of animal behavior in 3D is stereo vision \cite{cheng_obtaining_2018, wang_3d_2016-1, qian_feature_2017, cachat_three-dimensional_2011, cheng_3d_2016, viscido_individual_2004,straw_multi-camera_2010,muller_calibration_2014, stewart_novel_2015}.
3D tracking of zebrafish has been focused mainly on single subjects or small groups, as occlusion is a big hindrance for maintaining correct IDs due to their shoaling behavior \cite{miller_quantification_2007}.
Furthermore, the visual appearance of the fish can change dramatically depending on the position and posture, which makes re-identification more complex compared to 2D.

The Track3D module from the commercial EthoVision XT \cite{noldus_ethovision:_2001} is popular for tracking zebrafish in 3D, but is limited to a single individual \cite{cachat_three-dimensional_2011, stewart_novel_2015}.
An early semi-automatic 3D tracking system was developed by Viscido \etal \cite{viscido_individual_2004} to investigate the relationship between individual members of fish schools. 
Initial 2D tracks were generated by a nearest neighbor algorithm followed by a step allowing the user to adjust and correct the proposed 2D trajectories, and subsequently triangulated to reconstruct the 3D trajectories.

Qian \etal have worked extensively with tracking of zebrafish and have developed a 2D tracking system with a top-view camera using an augmented fast marching method (AFMM) \cite{qian_effective_2016} and the determinant of the Hessian \cite{qian_automatically_2014}. 
This was expanded to 3D tracking by extending the setup with a side-view camera. 
AFMM was utilized to generate a feature point based fish representation in each view followed by 2D tracklet construction based on motion constraints.
3D tracks were then constructed by associating the 2D tracklets with side-view detections using epipolar and motion consistency constraints \cite{qian_skeleton-based_2017}. Liu \etal \cite{Liu2019} extended this method to better handle occlusions based on a set of heuristic methods and the epipolar constraint.  
A third camera was added in \cite{qian_feature_2017}, and the feature point representation method was extended. 

\begin{table*}[!t]
\centering
\resizebox{!}{0.083\textheight}{%
\begin{tabular}{c| c c c c c c c c |c}
         & Trn2           & Trn5        & Val2  	  & Val5          & Tst1          & Tst2         & Tst5         & Tst10       & Total   \\ \hline
Length   & 120 s          & 15 s        & 30 s        & 15 s          & 15 s          & 15 s         & 15 s         & 15 s        & 240 s   \\
Frames   & 14,400         & 1,800       & 3,600       & 1,800         & 1,800         & 1,800        & 1,800        & 1,800       & 28,800  \\
BBs      & 28,800         & 9,000       & 7,200       & 9,000         & 1,800         & 3,600        & 9,000        & 18,000      & 86,400  \\ 
Points   & 28,800         & 9,000       & 7,200       & 9,000         & 1,800         & 3,600        & 9,000        & 18,000      & 86,400  \\ \hline \hline
OC       & 1.82 / 1.42    & 3.60 / 2.93 & 0.93 / 0.47 & 2.67 / 3.80   &  0.00 / 0.00   & 0.67 / 0.67  & 3.07 / 2.93  & 4.40 / 6.53 &        \\
OL       & 0.41 / 0.51    & 0.56 / 0.64 & 0.22 / 0.63 & 0.25 / 0.66   &  0.00 / 0.00   & 0.10 / 0.38  & 0.25 / 0.36  & 0.28 / 0.35 &        \\
TBO      & 0.69 / 0.89    & 1.00 / 1.21 & 1.79 / 3.20 & 1.64 / 0.73   & 15.00 / 15.00  & 2.41 / 2.18  & 1.38 / 1.28  & 1.86 / 1.40 &        \\
IBO      & 0.29 / 0.26    & 0.28 / 0.28 & 0.24 / 0.35 & 0.22 / 0.34   &  0.00 / 0.00   & 0.19 / 0.19  & 0.25 / 0.23  & 0.26 / 0.24 &        \\
\text{$\Psi$} & \textbf{0.26} & \textbf{0.50} & \textbf{0.03} & \textbf{0.63} & \textbf{0.00} & \textbf{0.01} & \textbf{0.16} & \textbf{0.28} &  
\end{tabular}%
}
\caption{Overview of the proposed dataset. OC, OL, TBO, and IBO are listed for the top- and front-view, respectively, and the number of fish is denoted in the sequence name. OC:~average amount of occlusions per second, OL:~average occlusion length in seconds, TBO:~average amount of seconds between occlusions, IBO:~intersection between occlusions, $\Psi$:~complexity measure based on OC, OL, TBO and IBO (see \equationref{psi}).}
\label{tab:dataset_specs}
\end{table*}

Cheng \etal \cite{cheng_obtaining_2018} utilized a similar three-camera setup, applying an iterative unsupervised learning method to train a CNN-based classifier to distinguish between the individual fish from a camera placed above the water tank.
The classifier was trained on the head region of the fish during periods when all fish were visible at the same time. 
By iteratively retraining the classifier, they were able to generate 2D tracks from the top-view and reconstruct the 3D tracklets based on detections from the two other side-view cameras under epipolar and motion constraints.

Wang \etal \cite{wang_3d_2016-1} also utilized a three-camera setup, using a Gaussian Mixture Model, a Gabor filter and an SVM-based method to detect the fish heads in the top- and side-views, respectively.
The top-view detections are associated into 2D tracklets based on a cross-correlation method and by applying a Kalman filter; near linear movement is achieved by a frame rate of 100 FPS. 
The 2D tracklets are then constructed into 3D tracklets by associating the side-view detections under epipolar and motion constraints. 
In \cite{Wang_LSTM_2D_2016}, Wang \etal proposed to model the top-view movement of the zebrafish through long short-term memory networks, which were used to improve the motion constraints in a new iteration of their 3D system \cite{Wang_LSTM_3D_2017}. 
Lastly, Wang \etal used a CNN for re-identification of zebrafish heads from the top-view \cite{wang_robust_2016}, although this has yet to be incorporated into a 3D tracking setup.
None of the methods are able to track multiple zebrafish in 3D for more than a few seconds without ID swaps; this is still a difficult and unsolved problem.

\textbf{Datasets.} As in other MOT challenges, there is a mutual agreement that occlusion is what makes 3D tracking of zebrafish difficult.
Nonetheless, only Wang \etal \cite{Wang_LSTM_3D_2017} describe their recordings based on occlusion frequency; however, they do not define how it is measured.
Qian \etal \cite{qian_feature_2017} indicate their complexity based on the amount of fish, but only four occlusion events occur during their 15 seconds demo video with ten fish.
For comparison, there are 66 occlusion events in our 15 seconds sequence with ten fish.

\section{Proposed Dataset}
The proposed 3D zebrafish dataset, 3D-ZeF, has been recorded from a top- and front-view perspective.
This approach was taken to minimize events of total occlusion typical for side-by-side binocular setups.
An example of the visual variation between the views is shown in \figref{examples} together with an illustration of the experimental setup.

\subsection{Experimental Setup}
The setup used to record the proposed dataset was built entirely from off-the-shelf hardware, whereas previous methods have used specialized camera equipment.
An illustration of the setup is shown in \figref{examples}.
The two light panels are IKEA FLOALT of size $30\times 30$ cm with a luminous flux of 670 lumen and a color temperature of 4000K.
The test tank is a standard glass aquarium of size $30\times 30\times 30$ cm with a water depth of 15 cm.
The top and front cameras are GoPro Hero 5 and GoPro Hero 7, respectively.
All the videos are recorded with a resolution of $2704\times 1520$, 60 FPS, 1/60 $s$ shutter speed, 400 ISO, and a linear field of view.
However, the fish tank does not take up the entire image, therefore, the effective region of interest is approximately $1200\times 1200$ and $1800\times 900$ for the top- and front-view, respectively.
Diffusion fabric was placed in front of the top light in order to reduce the amount of glare in the top-view.
Semi-transparent plastic was attached to three out of four of the window panes in order to reduce reflections. 
Furthermore, the front camera was placed orthogonally to the water level, which reduced reflections from the water surface.
Lastly, the pair-wise recordings have been manually synchronized using a flashing LED, which results in a worst case temporal shift of $\frac{1}{2\cdot \textsc{FPS}}$.

\subsection{Dataset Construction}
A total of eight sequences were recorded and divided into a training, validation, and test split.
Each sequence consists of a pair of temporally aligned top- and front-view videos and the specifications of the three splits are shown in \tableref{dataset_specs}. 
In order to avoid data leakage, each split contains a unique set of fish.
The training and validation set of fish were from the same cohort, whereas the fish in the test split were from a younger cohort.
Therefore, the test set differs from the training and validation set, as the fish are smaller and behave socially different.
This represent a real-life scenario where different cohorts need to be tracked, which has not generally been addressed within the field.

The zebrafish were manually bounding box and point annotated with consistent identity tags through all frames. 
The bounding boxes were tightly fitted to the visible parts of the zebrafish and the point annotations were centered on the head. 
If a set of fish touched, an occlusion tag was set for all involved bounding boxes. 
During occlusions, the bounding box was fitted to the visible parts of the fish and not where it was expected to be due to the extreme flexibility of the zebrafish. 
The pair-wise point annotations from the two views were triangulated into 3D positions using the method proposed by Pedersen \etal \cite{Pedersen_2018_CVPR_Workshops}. 
The fish head was approximated during occlusions to ensure continuous 3D tracks. 

It should be noted that the data was recorded in RGB.
Zebrafish can change their body pigmentation based on their environment, stress level, and more \cite{kalueff_towards_2013}. 
The changes in coloration can be important in behavioral studies and may even be valuable in solving the 3D tracking problem.

\subsection{Dataset Complexity}
Intuitively, a higher number of fish creates a more difficult tracking problem.
However, this is only true to some extent as the main complexity factor is the number and level of occlusions, which depends on a combination of the social activity and amount of space rather than the number of individuals. 
Therefore, we have defined a range of metrics based on occlusion events to describe the complexity of the proposed sequences.
An occlusion event is defined by a set of consecutive frames, where a fish is part of an occlusion.
The events are measured from the perspective of the fish; if two fish are part of an occlusion it counts as two events.

\begin{figure}[!t]
    \centering
    \includegraphics[width=0.45\textwidth]{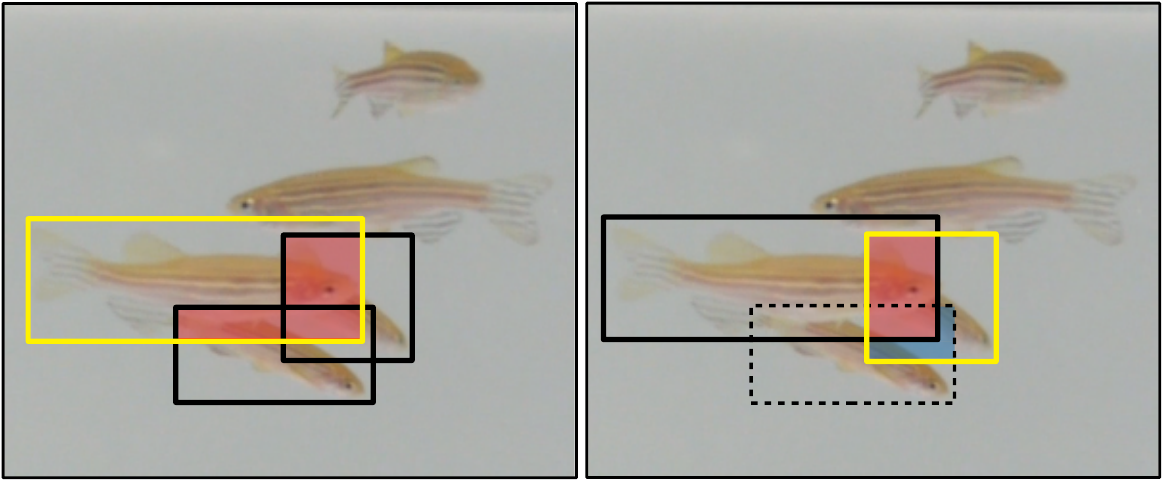}
    \caption{IBO seen from the perspective of two different individuals in the same frame. The targets are marked in yellow, the red area shows the intersection with a subject that is part of the same occlusion as the target, and the blue area shows the intersection with a subject that is not part of the same occlusion as the target.}
    \label{fig:io}
\end{figure}

The number of occlusion events indicates how often a fish is part of an occlusion, but, few long occlusions can be just as problematic as many short.
The length of the occlusions and time between them are, therefore, important to keep in mind when evaluating the complexity of a recording.
Due to our definition of occlusion events there are cases where fish are part of occlusions with only minor parts of their bodies.
Therefore, the intersection between occlusions is measured as an indication of the general intersection level.
The metrics that we provide as basis for the complexity level of our recordings are defined here:
\\\textbf{Occlusion Count (OC)}: the average number of occlusion events per second.
\\\textbf{Occlusion Length (OL)}: the average time in seconds of all occlusion events.
\\\textbf{Time Between Occlusions (TBO)}: the average time in seconds between occlusion events.
\\\textbf{Intersection Between Occlusions (IBO)}: a measure of how large a part of the fish that is part of an occlusion event.
The intersection in a frame, $f$, for fish $i$ is given by
\begin{equation}
    \textsc{IBO}_{i,f} = \frac{1}{|\text{bb}_i|}\sum_{j=1}^{n_\text{occ}}\text{bb}_i \cap \text{bb}_j,\ \ \ \text{for} \ \ j \neq i,
\end{equation}
where $n_{\text{occ}}$ is the number of fish in an occlusion event, and bb$_j$ is the set of pixel coordinates in the bounding box of fish $j$. 
IBO is measured across all bounding boxes with an occlusion tag in a given frame, even for subjects that are not part of the same occlusion.
Two examples are presented in \figref{io}, where the IBO$_{i,f}$ is calculated from the perspective of the targets enclosed in yellow.
The blue area in the second example, represents the intersection with a subject that is not part of the same occlusion as the target.
Additionally, the annotated bounding boxes enclose only the visible parts of the subjects.
Thus, the actual intersection between the subjects can be higher if a large part of a fish is hidden.
Nonetheless, the assumption is that a high IBO is an expression of heavy occlusion and vice versa.
The IBO measure presented in \tableref{dataset_specs} is an average between all fish in all frames.
A single complexity measure is calculated per sequence, by combining the four proposed metrics by
\begin{equation}\label{eq:psi}
    \Psi = \frac{1}{n}\sum_v^{\{\textsc{T,F}\}} \frac{\textsc{OC}_v \ \textsc{OL}_v \ \textsc{IBO}_v}{\text{TBO}_v},
\end{equation}
where $n$ is the number of camera views and subscript T and F denote the top- and front-view, respectively.
If a recording has no occlusions the complexity measure, $\Psi$, is zero; otherwise, the measure is in the interval $]0, \infty[$, where a larger value indicates a higher complexity.

\section{Method}
The pipeline of the proposed 3D tracker follows a modular tracking-reconstruction approach, where subjects are detected and tracked in each view before being triangulated and associated across views.
This allows us to use the temporal information of the tracklets in the two views in the 3D association step in opposition to a reconstruction-tracking approach, where detections are triangulated before tracks are generated.

\subsection{Object Detection in 2D}
A consistent 2D point is needed in each view in order to create 3D trajectories.
As the head is the only rigid part of the body the tracking point is chosen to be located between the eyes of the fish. 
We present two simple methods to find the head-point of the fish: a naive approach, that does not require training, and a CNN based approach.

\textbf{Naive:} A background image, $bg$, is initially estimated for each view by taking the median of $N_{bg}$ images sampled uniformly across the videos.
Subsequently, the background is subtracted by calculating the absolute difference image, $fg = |im-bg|$.
To locate the head of a fish in the top-view, the $fg$ is binarized using the intermodes bimodal threshold algorithm \cite{prewitt_analysis_1966}.
The skeletonization approach of Zhang and Suen \cite{zhang1984fast} is applied, and the endpoints are analyzed to locate the head of the fish.
In the front-view the $fg$ is binarized through the use of a histogram entropy thresholding method because the appearance of the fish cannot be approximated as bimodal.
The head point is estimated as being either the center of the blob or one of the middle edge points of the blob along the minor axis of the detected bounding box.
All three points are evaluated during the 3D reconstruction step, and the two points with the highest reprojection errors are discarded.

\textbf{FRCNN-H:} A Faster R-CNN \cite{ren2015faster} model has been trained for each view. The bounding boxes have been extracted from all the head-point annotations in the training sequences in order to train a head-detector model for each view.
The bounding boxes have static diameters of 25 and 50 pixels for the top-, and front-view, respectively.
The head-points are determined as the center of the detected bounding boxes which have a minimum confidence of $c$.

See the supplementary material for more detailed information on the detectors. 

\subsection{2D Tracklet Construction}
As zebrafish move erratically, it is difficult to set up a stable motion model. 
Therefore, we use a naive tracking-by-detection approach.
The tracking is done by constructing a distance matrix between the detections in a frame and the last detections of current tracklets. 
The matrix is solved as a global optimization problem using the Hungarian algorithm \cite{kuhn_hungarian_1955}. 
Tracklets are deliberately constructed in a conservative manner, where robustness is encouraged above length. 
A new detection is only assigned to a tracklet located within a  minimum distance, denoted $\delta_\text{T}$ and $\delta_\text{F}$, for the top and front view respectively. 
If a tracklet has not been assigned a detection within a given amount of time, $\tau_k$, the tracklet is terminated.

The $\ell_2$ distance between the head detections is used in both views for the FRCNN-H method.
However, the Mahalanobis distance between the center-of-mass is used for the front-view in the Naive method.
This is due to the elliptical form of the zebrafish body, which can be utilized by setting the covariance matrix of the blob as the Mahalanobis matrix; as the fish is more likely to move along the major axis than along the minor axis.

\subsection{2D Tracklet Association Between Views}
The 2D tracklets from each view are associated into 3D tracklets through a graph-based approach. 
All 2D tracklets with less than a given number of detections, $\alpha$, are removed in order to filter out noisy tracklets. 
The 3D calibration and triangulation method from Pedersen \etal \cite{Pedersen_2018_CVPR_Workshops} is used.

\subsubsection{Graph Construction}
A directed acyclic graph (DAG) is constructed. 
Every node represents a 3D tracklet and consists of two 2D tracklets; one from each camera view. 
Each edge associates nodes, where the 3D tracklet is based on the same 2D tracklet from one of the views.

\textbf{Create nodes}: The graph nodes are constructed by processing each top-view tracklet and identifying all temporally intersecting front-view tracklets as given by
\begin{equation}\label{eq:intersectingTracklets}
    I = F_{\text{T}}\cap F_{\text{F}},
\end{equation}
where $F_\text{T}$ and $F_\text{F}$ are the set of frames with detections in the top- and front-view tracklets, respectively, and $I$ is the set of frames with detections in both views.
If $I=\emptyset$, the node is not created.

\begin{figure}[!t]
    \centering
    \includegraphics[width=0.45\textwidth]{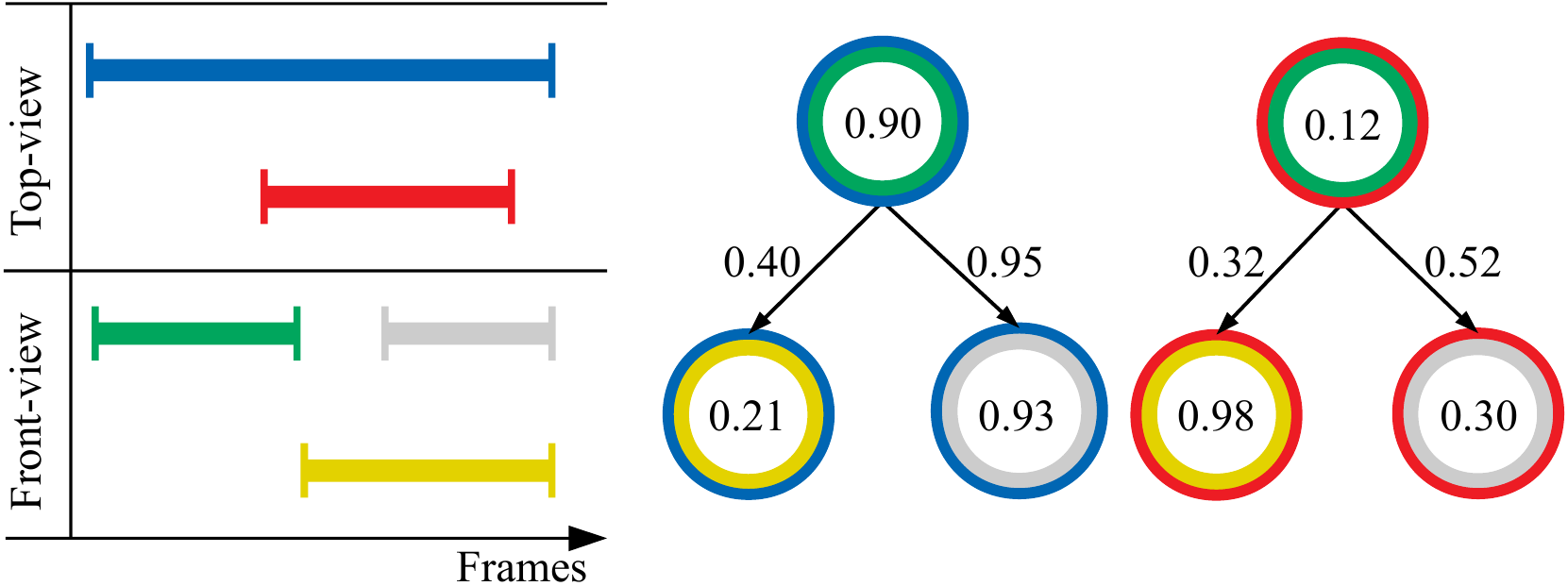}
    \caption{The colored lines represent 2D tracklets in each view, the node pairs are represented by the double-colored circles, and the edges of the DAG are shown by the arrows. The numbers represent example node and edge weights.}
    \label{fig:dag_construction}
\end{figure}

An example is presented in \figref{dag_construction}, where both the blue and red tracklets in the top-view intersects with the three tracklets in the front-view.
The outer and inner circles of the six nodes represent the top- and front-view tracklets, respectively.
The number inside the nodes indicates the node weight, which is calculated as follows.

For each intersecting frame in $I$, denoted $f$, the 2D tracklets are triangulated.
This results in a 3D point of the zebrafish head, $p_f$, with a reprojection error, $x_f$.
For the Naive method where the head is not directly detected in the front-view, the top-view 2D point is triangulated with the three estimated points to find the match resulting in the smallest reprojection error.
Therefore, $p_f$ represents the point with the smallest reprojection error.
To penalize large reprojection errors, the complimentary probability from the exponential cumulative distribution function (CDF), $\Phi$, is utilized.
The exponential CDF is chosen as it approximately models the reprojection error of the ground truth training data. The set of weights for all valid 3D points, $V$, can be described by the following set-builder notation
\begin{equation}\label{eq:frameWeight}
V = \{ 1-\Phi(x_f \ | \ \lambda_{err})\ \ | \ \ f \in I \wedge A(p_f)\},
\end{equation}
where $\lambda_{err}$ is the reciprocal of the mean of the training data reprojection error, and $A$ states whether $p_f$ is within the water tank.  
The per-frame weights in $V$ are combined into a single weight, $W$, for the entire node by
\begin{equation}\label{eq:nodeWeight}
    W = \text{median}(V) \frac{|V|}{|F_{\text{T}}\cup F_{\text{F}}|},
\end{equation}
and the node is added to the DAG given that $W\neq 0$.
This weighting scheme considers both the reprojection error and the ratio of frames with valid 3D points compared to the set of all frames $I$.
The median function is used instead of the mean function in order to counteract that a few extreme outliers skew the weight.

\textbf{Connect nodes:} The nodes in the DAG should be connected to all other nodes building on one of the same 2D tracklets, as long as the 2D tracklets in the other view do not overlap temporally, as illustrated in \figref{dag_construction}. 
This is done by constructing the set of node pairs, $P$, from the set of nodes in the DAG, $N$. 
Each element of $N$, denoted $n$, consists of the 2D tracklets, $t_\text{F}$ and $t_\text{T}$, the 3D tracklet, $t$, and the node weight, $W$.
Nodes $n_i$ and $n_j$ are considered a pair if $t_{i,\text{T}} = t_{j,\text{T}}$ or $t_{i,\text{F}} = t_{j,\text{F}}$, if the 2D tracklets in the other view do not temporally overlap, and if $t_i$ starts earlier in time than $t_j$.
This is necessary in order to avoid assigning multiple detections to the same frame.

This can be represented by the set-builder notation
\begin{equation}\label{eq:nodePairSet}
P = \{(n_i, n_j) \ | \ n_i,n_j \in N \wedge O(n_i, n_j) \wedge T(n_i, n_j)\},
\end{equation}
where $O$ assesses whether $t_i$ starts before $t_j$, and $T$ ensures that the 2D tracklets in $n_i$ and $n_j$ do not temporally overlap, where $n = \{t_\text{T}, t_\text{F}, t, W\}$.

For each node pair in $P$, the weight, $E$, of the directed edge from $n_i$ to $n_j$ is based on:
\begin{itemize}
    \item {$s$, the speed of the fish as it moves between the last detection in $t_i$ and the first detection in $t_j$.}
    \item {$t_d$, the temporal difference between $t_i$ and $t_j$.}
    \item {$W_i$ and $W_j$, the weights of the nodes.}
\end{itemize}
The edge weight is calculated as the complimentary probability of the CDF of the exponential distribution, $\Phi$. The exponential distribution is chosen as it approximately models that of the speed of the zebrafish. $E$ is calculated by
\begin{equation}\label{eq:edgeWeight}
E = (1-\Phi(s \ | \ \lambda_s)) e^{-\frac{t_d}{\tau_p}} (W_i + W_j),
\end{equation}
where $\tau_p$ is an empirically chosen value, and $\lambda_{s}$ is the reciprocal of the sum of the mean and standard deviation of the measured speed in the training data. 
In case a node is not present in any node pairs, the node will be assigned to the DAG, but it will have no edges. The DAG is therefore a disconnected graph.
\subsubsection{Graph Evaluation} 
\begin{figure}[!t]
    \centering
    \includegraphics[width=0.45\textwidth]{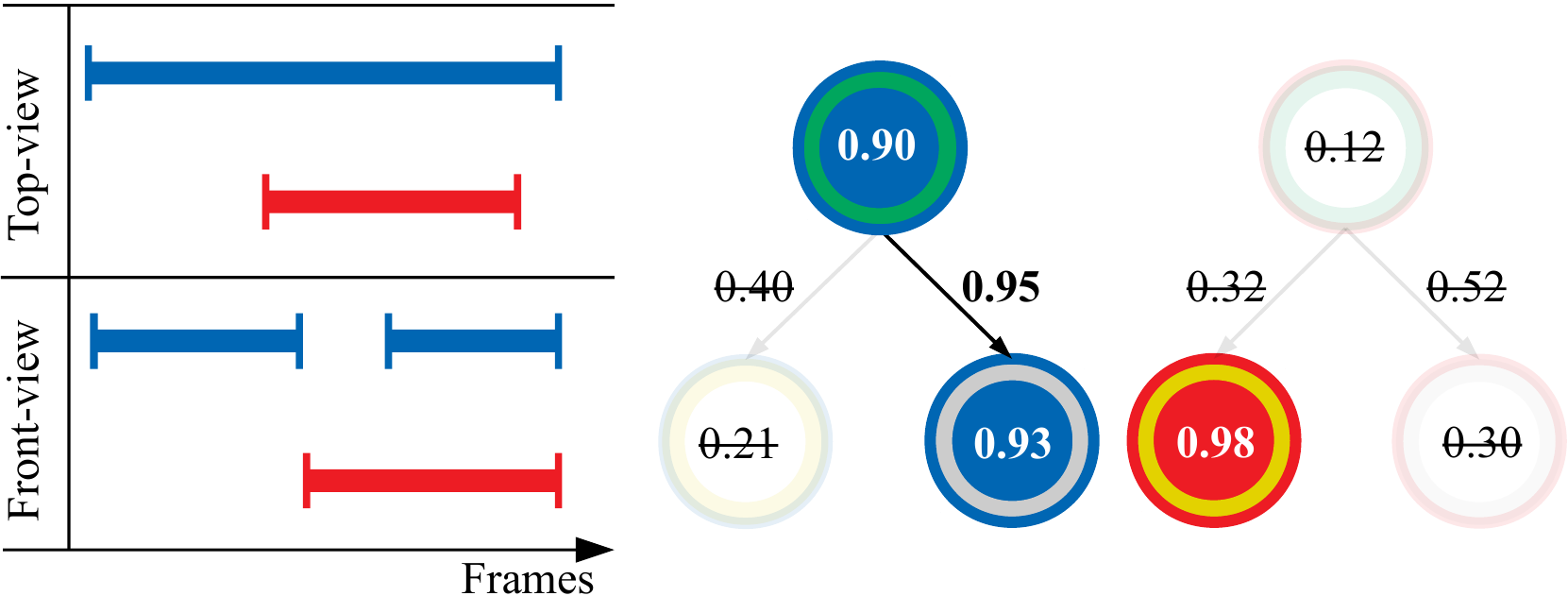}
    \caption{Graph evaluation based on the example from \figref{dag_construction}. The colored lines represent 2D tracklet pairs based on the chosen nodes in the graph; the transparent nodes are discarded.}
    \label{fig:dag_evaluation}
\end{figure}
The final 3D tracklets are extracted from the constructed DAG; this is done by recursively finding the longest path in the graph and storing the set of nodes as a single 3D tracklet.
The longest path is the path throughout the DAG, which gives the highest value when summing all nodes and edge weights in the path, see \figref{dag_evaluation}.
After extraction of a path, the used nodes, and all other nodes using the same 2D tracklets, are removed from the DAG.
This process is repeated until the DAG is empty.
In case a 2D tracklet in the 3D tracklet is missing a detection, the 3D position cannot be assigned, but the known information of the 2D tracklet is kept.
For the Naive method, the head position of the front-view 2D tracklet is determined by assigning the estimated point, which minimizes the $\ell_2$ distance to the head positions in the consecutive frame.

\subsection{3D Tracklet Association}
The final 3D tracks are constructed from the 3D tracklets in a greedy manner. 
A set of tracklets equal to the amount of fish present,  $N_{\text{fish}}$, is used as initial \textit{main tracklets}. 
The remaining tracklets, denoted \textit{gallery tracklets}, are assigned one by one to a single main tracklet, until no more tracklets can be assigned.

\subsubsection{Initial Tracklet Selection}
The set of $N_{\text{fish}}$ in the main tracks is selected by finding the stable tracklets that are temporally concurrent in time and span long time intervals. 
For each tracklet, the set of other temporally concurrent tracklets is considered.
In this set, all possible combinations of size $N_{\text{fish}}$ are investigated. If all tracklets in the set overlap temporally, the set is saved as a valid tracklet set.
The valid tracklet set with the highest median temporal overlap is used to construct $N_{\text{fish}}$ full 3D tracks.
This is done by using the greedy association scheme described in the following sections.
No 3D tracks are created if no valid combination of size $N_{\text{fish}}$ is identified.

\subsubsection{Greedy Association}
A greedy association algorithm is used when each gallery tracklet is associated with a single main tracklet.
The greedy part of the algorithm concerns the way that gallery tracklets are chosen; all gallery tracks are ranked in ascending order by the shortest temporal distance to any main tracklet. 
If the gallery tracklet overlaps temporally with all main tracklets, it is relegated to the end of the list. 
When the gallery tracklet has been associated with a main track, the remaining gallery tracks are re-ranked, and the process repeated.
In this way, the main tracklets are ``grown'' into full tracks. The gallery tracklet assignment is based on minimizing the cost of assignment. The cost is based on a set of distance measures, which are determined from two cases.

In the first case at least one main tracklet does not temporally overlap with the gallery tracklet. In this case, the association process is based on the spatio-temporal distances between the gallery tracklet and main tracklets. All temporally overlapping main tracklets are not considered.

\begin{figure}[t!]
    \centering
    \includegraphics[width=0.45\textwidth]{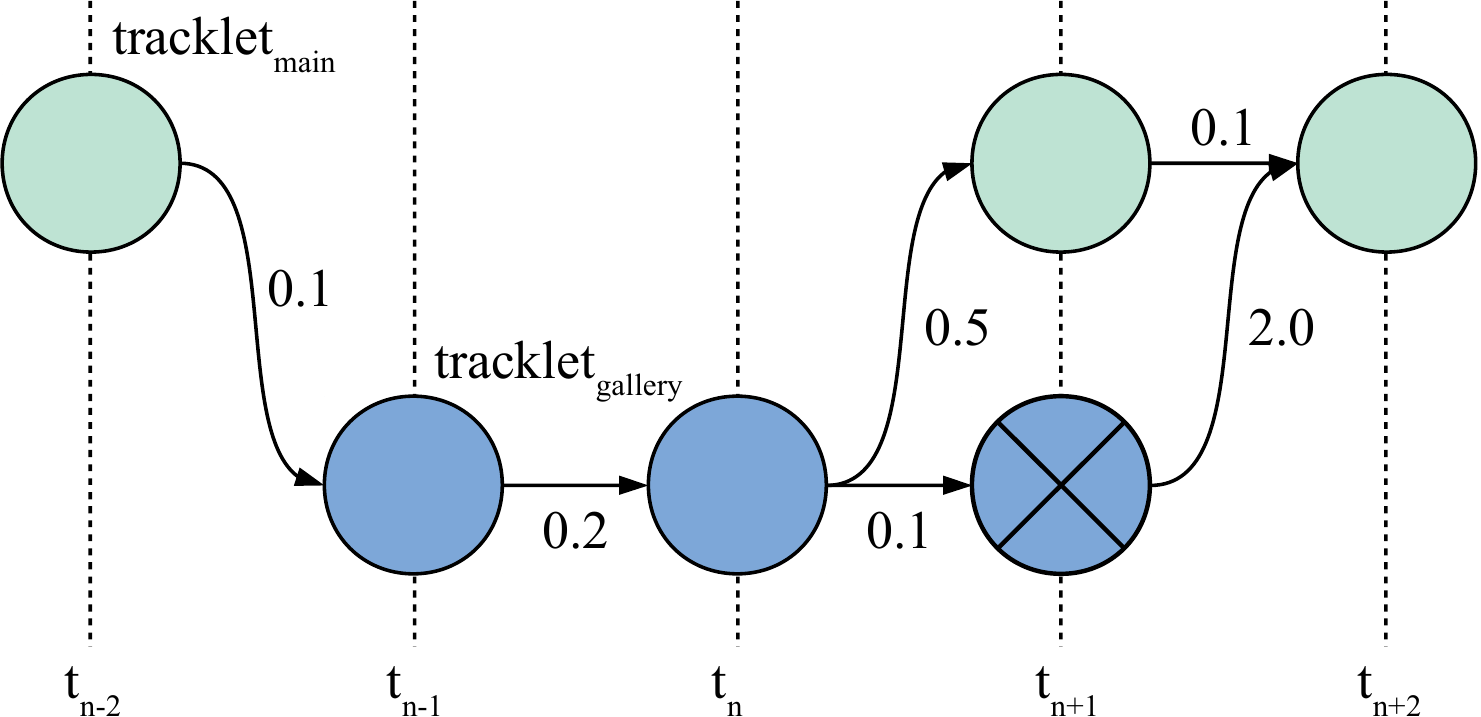}
    \caption{Example of internal spatio-temporal DAG, with the spatial distance between detections in the tracklets. The shortest path is found when switching from tracklet$_{\text{gallery}}$ to tracklet$_{\text{main}}$ in frame $t_{n+1}$.}
    \label{fig:interdistance}
\end{figure}
In the second case the gallery tracklet overlaps temporally with all main tracklets. As the spatio-temporal distances between the main and gallery tracklet is no longer measurable, a different set of distance values are used: The internal spatio-temporal distances, the amount of intersecting frames, \ie frames with a detection in both the main and gallery tracklets, and the ratio of intersecting frames compared to the total amount of detections in the gallery tracklet.
The internal spatio-temporal distances are determined through the construction of a DAG, where each node is a detection in a frame, and the edge weights are the spatial distances between the temporally previous nodes.
The final path is the one minimizing the spatial distance traveled.
An example of a graph is shown in \figref{interdistance}.
The distances are calculated as the mean of the values when the graph switches from a detection in the gallery tracklet to the main tracklet and vice versa.

\textbf{Association:} The distance measures are consolidated into a single assignment decision through a global cost scheme.
Each distance value is normalized across valid main tracklets into the range $[0;1]$ and sum to 1. The final cost of assigning the gallery tracklet to a main tracklet, is obtained by calculating the mean of the normalized distance values.
The gallery tracklet is associated with the main tracklet with the smallest cost, unless all main tracklet costs are located within a small margin, $\beta$, of each other, in which case the gallery tracklet is discarded. $\beta$ directly enforces a margin of confidence in the assignment, in order to not assign a gallery traklet based on inconclusive cost values.

\section{Evaluation}
The metrics used in the MOT challenges \cite{MOTChallenge2015, MOT16, MOT19_CVPR} and the Mean Time Between Failures (MTBF) proposed by Carr and Collins \cite{MTBF} are utilized to measure the performance of the system on the proposed dataset.
The MOT challenge metrics consist of the CLEAR MOT metrics \cite{CLEARMOT}, the mostly tracked/lost metrics \cite{MTML_Metrics}, and the identification-based metrics \cite{Ristani_id_metrics}.

The final 3D tracks are evaluated based on a subset of the MOT challenge metrics and the monotonic MTBF metric.
The primary metric used is the multiple object tracking (MOTA) metric.
The detected and ground truth tracklets are compared using the detected and annotated head points.
A detection is only associated with a ground truth tracklet if it is within a distance of 0.5 cm.
\begin{figure}[!t]
    \centering
    \includegraphics[width=0.50\textwidth]{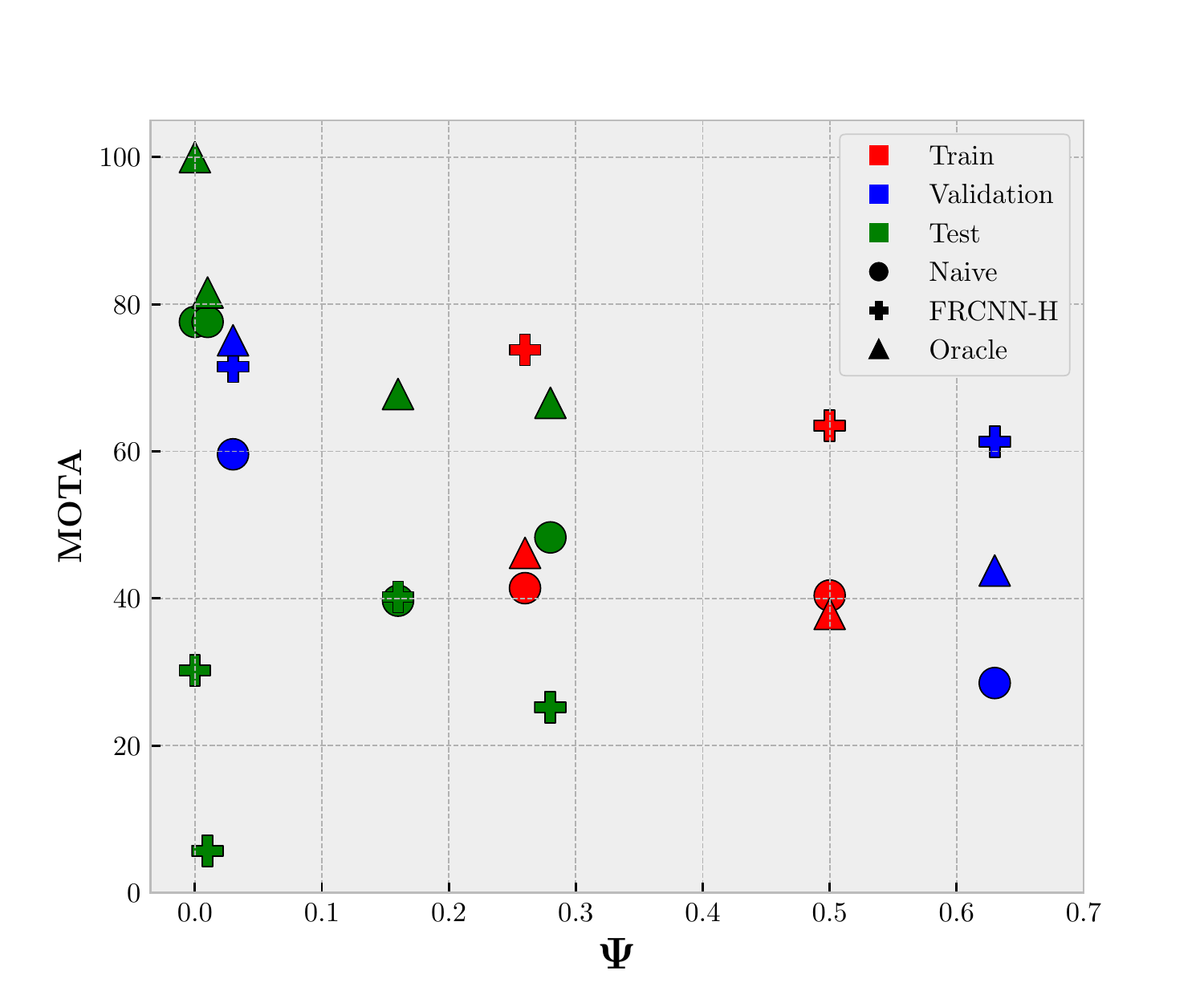}
    \caption{MOTA compared to the dataset complexity, $\Psi$, for all sequences in the dataset.}
    \label{fig:complexity}
\end{figure}
The performance of the system is evaluated with two different detection modules: Naive, and FRCNN-H.
The results are compared with a hypothetical tracker, called Oracle, which tracks perfectly at all times except during occlusions.
This provides an upper bound on the performance if occlusions are not handled in any way.
The full set of metrics, system parameters, and results can be found in the supplementary material. 

Results for all sequences compared to data complexity is shown in \figref{complexity}, and metrics for the test sequences are shown in \tableref{trackMeasures}. 
It is clear that the FRCNN-H outperforms the Naive method on the training and validation splits; it even outperforms the Oracle tracker in three out of four cases.
This is likely due to the method being able to detect some of the fish heads during occlusions. 
However, the superior performance is only seen on the two splits where the fish are from the same cohort.
On the test set the FRCNN-H fails to generalize, whereas the Naive method still manages to track the fish.

It should be noted that the poor performance of the Naive method on Tst1, is suspected to be due many short tracks from erratic movement, which the pipeline with the used parameter settings does not handle well.

\subsection{Comparison with Other Methods}
It has not been possible to make a fair comparison with the other 3D zebrafish tracking methods mentioned in \sectionref{related_work}.
Previous systems have been analyzed in terms of ID swaps, fragments, precision, and recall for the generated 2D and 3D tracks. 
However, there is no exact description of how these metrics are calculated. 
The evaluation protocol is further limited by not including a statement on the maximum allowed distance between estimated and ground truth tracks leading to uncertainty on the accuracy of the metrics.

Furthermore, the evaluated sequences are not described in terms of complexity, even though occlusion is repeatedly stated as a major hindrance in 3D zebrafish tracking. 
The only common complexity indication of the datasets is the number of fish, even though it is not representative.
An example of this is the tracking demo video of Qian \etal \cite{qian_effective_2016} with ten fish and only four occlusion events during 15 seconds.
Wang \etal \cite{wang_3d_2016-1} describes their dataset on basis of an occlusion probability but do not explain how it is measured.

There are currently no publicly available annotated data and the previous systems are evaluated on seemingly simplified cases of the problem.
Furthermore, the used evaluation protocols are lacking details in such a manner that it is not possible to determine under which conditions the metrics have been calculated. 
This, along with inaccessible codebases, severely limits the reproducibility of the results, and makes it impossible to ensure identical evaluation procedures. 
Therefore, it simply does not make sense to compare the proposed system to the other methods under the current circumstances.
\begin{table}[t!]
\centering
\resizebox{\linewidth}{!}{%
\begin{tabular}{clcccccccc}
\hline
& Method & MOTA $\uparrow$  & MT $\uparrow$ & ML $\downarrow$ & ID Sw. $\downarrow$ & Frag. $\downarrow$ & $\textrm{MTBF}_{m}$ $\uparrow$ \\ \hline
\multirow{3}{*}{\STAB{\rotatebox[origin=c]{90}{Tst1}}} 
 & Naive            &  77.6\% & 1 & 0 & 0 & 28   & 12.5 \\
 & FRCNN-H          &  30.2\% & 0 & 0 & 0 & 15   & 8.212 \\
 & Oracle           & 100.0\% & 1 & 0 & 0 & 0    & 900 \\ \hline 
\multirow{3}{*}{\STAB{\rotatebox[origin=c]{90}{Tst2}}} 
 & Naive            & 77.6\%  & 1 & 0 & 0 & 44   & 15.856 \\
 & FRCNN-H          &  5.7\%  & 0 & 2 & 2 & 17   & 2.641 \\
 & Oracle           & 81.6\%  & 2 & 0 & 0 & 25   & 27.396 \\ \hline 
\multirow{3}{*}{\STAB{\rotatebox[origin=c]{90}{Tst5}}} 
 & Naive            & 39.7\%  & 0 & 0 & 7 & 185  & 6.249 \\
 & FRCNN-H          & 40.2\%  & 0 & 0 & 7 & 115  & 7.577 \\
 & Oracle           & 67.8\%  & 1 & 0 & 0 & 50   & 28.112 \\ \hline 
\multirow{3}{*}{\STAB{\rotatebox[origin=c]{90}{Tst10}}} 
 & Naive            & 48.3\%  & 0 & 0 & 11 & 268 &  9.075 \\
 & FRCNN-H          & 25.2\%  & 0 & 3 & 32 & 225 & 4.904 \\
 & Oracle           & 66.6\%  & 1 & 10 & 0 & 119 & 23.105 \\ \hline 
\end{tabular}%
}
\caption{Evaluation of 3D tracks on test split. The arrows indicate whether higher or lower values are better. MOTA: Multiple Object Tracking Accuracy, MT: Mostly tracked, ML: Mostly lost, ID Sw.: Number of identity swaps, Frag.: Number of fragments, $\textrm{MTBF}_{m}$: Monotonic MTBF.}
\label{tab:trackMeasures}
\end{table}

\section{Conclusion}
Zebrafish is an increasingly popular animal model and behavioral analysis plays a major role in neuroscientific and biological research.
However, it is tedious and subjective to manually describe the complex 3D motion of zebrafish.
Therefore, 3D zebrafish tracking systems are critically needed to conduct accurate experiments on a grand scale.
The significant development experienced in other fields of MOT has not yet translated to 3D zebrafish tracking.
The main reason being that no dataset has been made publicly available with ground truth annotations.
Therefore, we present the first publicly available RGB 3D zebrafish tracking dataset called 3D-ZeF.

3D-ZeF consists of eight stereo sequences with highly social and similarly looking subjects demonstrating complex and erratic motion patterns in three dimensions that are not seen in common MOT challenges. 
A complexity measure based on the level of occlusions has been provided for each sequence to make them comparable to future related datasets.
The proposed dataset is annotated with 86,400 bounding boxes and points; the latter used for estimating ground truth 3D tracks based on the head position of the fish.
Different cohorts of zebrafish are used in the training, validation, and test splits to avoid data leakage; a problem that has never been addressed within the field.

The proposed Naive method scores a MOTA between 25\% and 80\% across the entire dataset, which correlates well with the complexity measure of the recordings.
The open-source modular based system provides a baseline and stepping stone for further development within the field of 3D zebrafish tracking and understanding.

{\small
\bibliographystyle{ieeetr}
\bibliography{zebrafish}

\begin{thebibliography}{10}

\bibitem{eisen_zebrafish_1996}
J.~S. Eisen, ``Zebrafish make a big splash,'' {\em Cell}, vol.~87,
  pp.~969--977, Dec. 1996.

\bibitem{haffter_identification_1996}
P.~Haffter, M.~Granato, M.~Brand, M.~Mullins, M.~Hammerschmidt, D.~Kane,
  J.~Odenthal, F.~van Eeden, Y.~Jiang, C.~Heisenberg, R.~Kelsh,
  M.~Furutani-Seiki, E.~Vogelsang, D.~Beuchle, U.~Schach, C.~Fabian, and
  C.~Nusslein-Volhard, ``The identification of genes with unique and essential
  functions in the development of the zebrafish, {Danio} rerio,'' {\em
  Development}, vol.~123, no.~1, pp.~1--36, 1996.

\bibitem{collier_zebrafish_2014}
A.~D. Collier, K.~M. Khan, E.~M. Caramillo, R.~S. Mohn, and D.~J. Echevarria,
  ``Zebrafish and conditioned place preference: {A} translational model of drug
  reward,'' {\em Progress in Neuro-Psychopharmacology and Biological
  Psychiatry}, vol.~55, pp.~16--25, Dec. 2014.

\bibitem{lin_zebrafish_2016}
C.-Y. Lin, C.-Y. Chiang, and H.-J. Tsai, ``Zebrafish and {Medaka}: new model
  organisms for modern biomedical research,'' {\em Journal of Biomedical
  Science}, vol.~23, Jan. 2016.

\bibitem{soares_using_2018}
M.~C. Soares, S.~C. Cardoso, T.~d.~S. Carvalho, and C.~Maximino, ``Using model
  fish to study the biological mechanisms of cooperative behaviour: {A} future
  for translational research concerning social anxiety disorders?,'' {\em
  Progress in Neuro-Psychopharmacology and Biological Psychiatry}, vol.~82,
  pp.~205--215, Mar. 2018.

\bibitem{kalueff_zebrafish_2014}
A.~V. Kalueff, A.~M. Stewart, and R.~Gerlai, ``Zebrafish as an emerging model
  for studying complex brain disorders,'' {\em Trends in Pharmacological
  Sciences}, vol.~35, pp.~63--75, Feb. 2014.

\bibitem{meshalkina_zebrafish_2018}
D.~A. Meshalkina, M.~N. Kizlyk, E.~V. Kysil, A.~D. Collier, D.~J. Echevarria,
  M.~S. Abreu, L.~J.~G. Barcellos, C.~Song, J.~E. Warnick, E.~J. Kyzar, and
  A.~V. Kalueff, ``Zebrafish models of autism spectrum disorder,'' {\em
  Experimental Neurology}, vol.~299, pp.~207--216, Jan. 2018.

\bibitem{khan_zebrafish_2017}
K.~M. Khan, A.~D. Collier, D.~A. Meshalkina, E.~V. Kysil, S.~L. Khatsko,
  T.~Kolesnikova, Y.~Y. Morzherin, J.~E. Warnick, A.~V. Kalueff, and D.~J.
  Echevarria, ``Zebrafish models in neuropsychopharmacology and {CNS} drug
  discovery,'' {\em British Journal of Pharmacology}, vol.~174, no.~13,
  pp.~1925--1944, 2017.

\bibitem{li_dominant_1997}
L.~Li and J.~E. Dowling, ``A dominant form of inherited retinal degeneration
  caused by a non-photoreceptor cell-specific mutation,'' {\em Proceedings of
  the National Academy of Sciences}, vol.~94, pp.~11645--11650, Oct. 1997.

\bibitem{muller_swimming_2004}
U.~K. Muller, ``Swimming of larval zebrafish: ontogeny of body waves and
  implications for locomotory development,'' {\em Journal of Experimental
  Biology}, vol.~207, pp.~853--868, Feb. 2004.

\bibitem{mcelligott_prey_2005}
M.~B. McElligott and D.~M. O'Malley, ``Prey tracking by larval zebrafish: Axial
  kinematics and visual control,'' {\em Brain, Behavior and Evolution},
  vol.~66, no.~3, pp.~177--196, 2005.

\bibitem{fontaine_automated_2008}
E.~Fontaine, D.~Lentink, S.~Kranenbarg, U.~K. Müller, J.~L. van Leeuwen, A.~H.
  Barr, and J.~W. Burdick, ``Automated visual tracking for studying the
  ontogeny of zebrafish swimming,'' {\em Journal of Experimental Biology},
  vol.~211, no.~8, pp.~1305--1316, 2008.

\bibitem{risse_fimtrack:_2017}
B.~Risse, D.~Berh, N.~Otto, C.~Klämbt, and X.~Jiang, ``{FIMTrack}: {An} open
  source tracking and locomotion analysis software for small animals,'' {\em
  PLOS Computational Biology}, vol.~13, no.~5, pp.~1--15, 2017.

\bibitem{ohayon_automated_2013}
S.~Ohayon, O.~Avni, A.~L. Taylor, P.~Perona, and S.~E.~R. Egnor, ``Automated
  multi-day tracking of marked mice for the analysis of social behaviour,''
  {\em Journal of Neuroscience Methods}, vol.~219, no.~1, pp.~10 -- 19, 2013.

\bibitem{qian_automatically_2014}
Z.-M. Qian, X.~E. Cheng, and Y.~Q. Chen, ``Automatically detect and track
  multiple fish swimming in shallow water with frequent occlusion,'' {\em PLOS
  ONE}, vol.~9, no.~9, pp.~1--12, 2014.

\bibitem{feijo_algorithm_2018}
G.~d.~O. Feijó, V.~A. Sangalli, I.~N. L.~d. Silva, and M.~S. Pinho, ``An
  algorithm to track laboratory zebrafish shoals,'' {\em Computers in Biology
  and Medicine}, vol.~96, pp.~79 -- 90, 2018.

\bibitem{franco-restrepo_review_2019}
J.~E. Franco-Restrepo, D.~A. Forero, and R.~A. Vargas, ``A review of freely
  available, open-source software for the automated analysis of the behavior of
  adult zebrafish,'' {\em Zebrafish}, vol.~16, June 2019.

\bibitem{romero-ferrero_idtracker.ai:_2019}
F.~Romero-Ferrero, M.~G. Bergomi, R.~C. Hinz, F.~J.~H. Heras, and G.~G.~d.
  Polavieja, ``idtracker.ai: tracking all individuals in small or large
  collectives of unmarked animals,'' {\em Nature Methods}, vol.~16,
  pp.~179--182, Jan. 2019.

\bibitem{noldus_ethovision:_2001}
L.~P. J.~J. Noldus, A.~J. Spink, and R.~A.~J. Tegelenbosch, ``{EthoVision}: {A}
  versatile video tracking system for automation of behavioral experiments,''
  {\em Behavior Research Methods, Instruments, \& Computers}, vol.~33,
  pp.~398--414, Aug. 2001.

\bibitem{systems_lolitrack_2018}
{Loligo Systems}, ``{LoliTrack} v.4.''
\newblock \url{https://www.loligosystems.com/lolitrack-v-4}.

\bibitem{viewpoint_zebralab_nodate}
{ViewPoint}, ``{ZebraLab}.''
\newblock \url{http://www.viewpoint.fr/en/p/software/zebralab}.

\bibitem{systems_videomot2_2014}
{TSE Systems}, ``{VideoMot2} - versatile video tracking system.''
\newblock \url{https://www.tse-systems.com/product-details/videomot}.

\bibitem{kalueff_towards_2013}
A.~V. Kalueff, M.~Gebhardt, A.~M. Stewart, J.~M. Cachat, M.~Brimmer, J.~S.
  Chawla, C.~Craddock, E.~J. Kyzar, A.~Roth, S.~Landsman, S.~Gaikwad,
  K.~Robinson, E.~Baatrup, K.~Tierney, A.~Shamchuk, W.~Norton, N.~Miller,
  T.~Nicolson, O.~Braubach, C.~P. Gilman, J.~Pittman, D.~B. Rosemberg,
  R.~Gerlai, D.~Echevarria, E.~Lamb, S.~C.~F. Neuhauss, W.~Weng, L.~Bally-Cuif,
  H.~Schneider, and t.~Z. Neuros, ``Towards a comprehensive catalog of
  zebrafish behavior 1.0 and beyond,'' {\em Zebrafish}, vol.~10, pp.~70--86,
  Mar. 2013.

\bibitem{cachat_video-aided_2011}
J.~M. Cachat, P.~R. Canavello, S.~I. Elkhayat, B.~K. Bartels, P.~C. Hart, M.~F.
  Elegante, E.~C. Beeson, A.~L. Laffoon, W.~A. Haymore, D.~H. Tien, A.~K. Tien,
  S.~Mohnot, and A.~V. Kalueff, ``Video-aided analysis of zebrafish locomotion
  and anxiety-related behavioral responses,'' in {\em Zebrafish
  {Neurobehavioral} {Protocols}} (A.~V. Kalueff and J.~M. Cachat, eds.),
  pp.~1--14, Totowa, NJ: Humana Press, 2011.

\bibitem{macri_three-dimensional_2017}
S.~Macrì, D.~Neri, T.~Ruberto, V.~Mwaffo, S.~Butail, and M.~Porfiri,
  ``Three-dimensional scoring of zebrafish behavior unveils biological
  phenomena hidden by two-dimensional analyses,'' {\em Scientific Reports},
  vol.~7, May 2017.

\bibitem{miller_quantification_2007}
N.~Miller and R.~Gerlai, ``Quantification of shoaling behaviour in zebrafish
  ({Danio} rerio),'' {\em Behavioural Brain Research}, vol.~184, pp.~157--166,
  Dec. 2007.

\bibitem{alzubi_real-time_2015}
H.~AlZu'bi, W.~Al-Nuaimy, J.~Buckley, L.~Sneddon, and {Iain Young}, ``Real-time
  {3D} fish tracking and behaviour analysis,'' in {\em 2015 {IEEE} {Jordan}
  {Conference} on {Applied} {Electrical} {Engineering} and {Computing}
  {Technologies} ({AEECT})}, pp.~1--5, Nov. 2015.

\bibitem{cheng_obtaining_2018}
X.~E. Cheng, S.~S. Du, H.~Y. Li, J.~F. Hu, and M.~L. Chen, ``Obtaining
  three-dimensional trajectory of multiple fish in water tank via video
  tracking,'' {\em Multimedia Tools and Applications}, vol.~77,
  pp.~24499--24519, Feb. 2018.

\bibitem{qian_skeleton-based_2017}
Z.~Qian, M.~Shi, M.~Wang, and T.~Cun, ``Skeleton-based {3D} tracking of
  multiple fish from two orthogonal views,'' in {\em Communications in
  {Computer} and {Information} {Science}}, pp.~25--36, Springer Singapore,
  2017.

\bibitem{wang_3d_2016-1}
S.~H. Wang, X.~Liu, J.~Zhao, Y.~Liu, and Y.~Q. Chen, ``{3D} tracking swimming
  fish school using a master view tracking first strategy,'' in {\em 2016
  {IEEE} {International} {Conference} on {Bioinformatics} and {Biomedicine}
  ({BIBM})}, IEEE, Dec. 2016.

\bibitem{qian_feature_2017}
Z.-M. Qian and Y.~Q. Chen, ``Feature point based {3D} tracking of multiple fish
  from multi-view images,'' {\em PLOS ONE}, vol.~12, no.~6, pp.~1--18, 2017.

\bibitem{audira_simple_2018}
G.~Audira, B.~Sampurna, S.~Juniardi, S.-T. Liang, Y.-H. Lai, and C.-D. Hsiao,
  ``A simple setup to perform {3D} locomotion tracking in zebrafish by using a
  single camera,'' {\em Inventions}, vol.~3, p.~11, Feb. 2018.

\bibitem{xiao_research_2016}
G.~Xiao, W.-K. Fan, J.-F. Mao, Z.-B. Cheng, D.-H. Zhong, and Y.~Li, ``Research
  of the fish tracking method with occlusion based on monocular stereo
  vision,'' in {\em 2016 {International} {Conference} on {Information} {System}
  and {Artificial} {Intelligence} ({ISAI})}, IEEE, June 2016.

\bibitem{Menze2015ISA}
M.~Menze, C.~Heipke, and A.~Geiger, ``Joint {3D} estimation of vehicles and
  scene flow,'' in {\em ISPRS Workshop on Image Sequence Analysis (ISA)}, 2015.

\bibitem{Voigtlaender2019CVPR}
P.~Voigtlaender, M.~Krause, A.~Osep, J.~Luiten, B.~B.~G. Sekar, A.~Geiger, and
  B.~Leibe, ``{MOTS:} multi-object tracking and segmentation,'' in {\em
  Conference on Computer Vision and Pattern Recognition (CVPR)}, 2019.

\bibitem{caesar2019nuscenes}
H.~Caesar, V.~Bankiti, A.~H. Lang, S.~Vora, V.~E. Liong, Q.~Xu, A.~Krishnan,
  Y.~Pan, G.~Baldan, and O.~Beijbom, ``{nuScenes:} a multimodal dataset for
  autonomous driving,'' {\em arXiv preprint arXiv:1903.11027}, 2019.

\bibitem{waymo_open_dataset}
``Waymo open dataset: An autonomous driving dataset,'' 2019.
\newblock \url{https://www.waymo.com/open}.

\bibitem{chang2019argoverse}
M.-F. Chang, J.~Lambert, P.~Sangkloy, J.~Singh, S.~Bak, A.~Hartnett, D.~Wang,
  P.~Carr, S.~Lucey, D.~Ramanan, {\em et~al.}, ``Argoverse: {3D} tracking and
  forecasting with rich maps,'' in {\em Proceedings of the IEEE Conference on
  Computer Vision and Pattern Recognition}, pp.~8748--8757, 2019.

\bibitem{song2013tracking}
S.~Song and J.~Xiao, ``Tracking revisited using {RGBD} camera: Unified
  benchmark and baselines,'' in {\em Proceedings of the IEEE international
  conference on computer vision}, pp.~233--240, 2013.

\bibitem{lukezic2019cdtb}
A.~Lukezic, U.~Kart, J.~Kapyla, A.~Durmush, J.-K. Kamarainen, J.~Matas, and
  M.~Kristan, ``Cdtb: A color and depth visual object tracking dataset and
  benchmark,'' in {\em Proceedings of the IEEE International Conference on
  Computer Vision}, pp.~10013--10022, 2019.

\bibitem{MOTChallenge2015}
L.~Leal-Taix\'{e}, A.~Milan, I.~Reid, S.~Roth, and K.~Schindler,
  ``{MOTC}hallenge 2015: {T}owards a benchmark for multi-target tracking,''
  {\em arXiv:1504.01942 [cs]}, Apr. 2015.
\newblock arXiv: 1504.01942.

\bibitem{MOT16}
A.~Milan, L.~Leal-Taix\'{e}, I.~Reid, S.~Roth, and K.~Schindler, ``{MOT}16: {A}
  benchmark for multi-object tracking,'' {\em arXiv:1603.00831 [cs]}, Mar.
  2016.
\newblock arXiv: 1603.00831.

\bibitem{MOT19_CVPR}
P.~Dendorfer, H.~Rezatofighi, A.~Milan, J.~Shi, D.~Cremers, I.~Reid, S.~Roth,
  K.~Schindler, and L.~Leal-Taix\'{e}, ``{CVPR19} tracking and detection
  challenge: {H}ow crowded can it get?,'' {\em arXiv:1906.04567 [cs]}, June
  2019.
\newblock arXiv: 1906.04567.

\bibitem{UA-DETRAC_2017}
S.~{Lyu}, M.~{Chang}, D.~{Du}, L.~{Wen}, H.~{Qi}, Y.~{Li}, Y.~{Wei}, L.~{Ke},
  T.~{Hu}, M.~{Del Coco}, P.~{Carcagnì}, D.~{Anisimov}, E.~{Bochinski},
  F.~{Galasso}, F.~{Bunyak}, G.~{Han}, H.~{Ye}, H.~{Wang}, K.~{Palaniappan},
  K.~{Ozcan}, L.~{Wang}, L.~{Wang}, M.~{Lauer}, N.~{Watcharapinchai},
  N.~{Song}, N.~M. {Al-Shakarji}, S.~{Wang}, S.~{Amin}, S.~{Rujikietgumjorn},
  T.~{Khanova}, T.~{Sikora}, T.~{Kutschbach}, V.~{Eiselein}, W.~{Tian},
  X.~{Xue}, X.~{Yu}, Y.~{Lu}, Y.~{Zheng}, Y.~{Huang}, and Y.~{Zhang},
  ``{UA-DETRAC} 2017: Report of {AVSS2017} {IWT4S} challenge on advanced
  traffic monitoring,'' in {\em 2017 14th IEEE International Conference on
  Advanced Video and Signal Based Surveillance (AVSS)}, pp.~1--7, Aug 2017.

\bibitem{UA-DETRAC_2020}
L.~Wen, D.~Du, Z.~Cai, Z.~Lei, M.-C. Chang, H.~Qi, J.~Lim, M.-H. Yang, and
  S.~Lyu, ``Ua-detrac: A new benchmark and protocol for multi-object detection
  and tracking,'' {\em Computer Vision and Image Understanding}, vol.~193,
  p.~102907, 2020.

\bibitem{DAN_2019}
S.~{Sun}, N.~{Akhtar}, H.~{Song}, A.~S. {Mian}, and M.~{Shah}, ``Deep affinity
  network for multiple object tracking,'' {\em IEEE Transactions on Pattern
  Analysis and Machine Intelligence}, pp.~1--1, 2019.

\bibitem{IOUTrack_2018}
E.~{Bochinski}, T.~{Senst}, and T.~{Sikora}, ``Extending {IOU} based
  multi-object tracking by visual information,'' in {\em 2018 15th IEEE
  International Conference on Advanced Video and Signal Based Surveillance
  (AVSS)}, pp.~1--6, Nov 2018.

\bibitem{Bergmann_2019_ICCV}
P.~Bergmann, T.~Meinhardt, and L.~Leal-Taixe, ``Tracking without bells and
  whistles,'' in {\em The IEEE International Conference on Computer Vision
  (ICCV)}, October 2019.

\bibitem{perez-escudero_idtracker:_2014}
A.~Perez-Escudero, J.~Vicente-Page, R.~C. Hinz, S.~Arganda, and G.~G.
  de~Polavieja, ``{idTracker}: tracking individuals in a group by automatic
  identification of unmarked animals,'' {\em Nature Methods}, vol.~11, pp.~743
  -- 748, 2014.

\bibitem{sridhar_tracktor:_2019}
V.~H. Sridhar, D.~G. Roche, and S.~Gingins, ``Tracktor: Image-based automated
  tracking of animal movement and behaviour,'' {\em Methods in Ecology and
  Evolution}, vol.~10, pp.~815--820, Mar. 2019.

\bibitem{monck_biotracker:_2018}
H.~J. Mönck, A.~Jörg, T.~v. Falkenhausen, J.~Tanke, B.~Wild, D.~Dormagen,
  J.~Piotrowski, C.~Winklmayr, D.~Bierbach, and T.~Landgraf, ``{BioTracker:} an
  open-source computer vision framework for visual animal tracking,'' {\em
  CoRR}, vol.~abs/1803.07985, 2018.

\bibitem{rodriguez_toxtrac_2017}
A.~Rodriguez, H.~Zhang, J.~Klaminder, T.~Brodin, P.~L. Andersson, and
  M.~Andersson, ``{ToxTrac:} a fast and robust software for tracking
  organisms,'' {\em Methods in Ecology and Evolution}, vol.~9, pp.~460--464,
  Sept. 2017.

\bibitem{harmer_pathtrackr_2019}
A.~M.~T. Harmer and D.~B. Thomas, ``pathtrackr: An r package for video tracking
  and analysing animal movement,'' {\em Methods in Ecology and Evolution}, May
  2019.

\bibitem{liu_tracking_2019}
X.~Liu, P.~R. Zhu, Y.~Liu, and J.~W. Zhao, ``Tracking full-body motion of
  multiple fish with midline subspace constrained multicue optimization,'' {\em
  Scientific Programming}, vol.~2019, pp.~1--7, June 2019.

\bibitem{zhu_catadioptric_2007}
L.~Zhu and W.~Weng, ``Catadioptric stereo-vision system for the real-time
  monitoring of {3D} behavior in aquatic animals,'' {\em Physiology \&
  Behavior}, vol.~91, no.~1, pp.~106 -- 119, 2007.

\bibitem{cachat_three-dimensional_2011}
J.~Cachat, A.~Stewart, E.~Utterback, P.~Hart, S.~Gaikwad, K.~Wong, E.~Kyzar,
  N.~Wu, and A.~V. Kalueff, ``Three-dimensional neurophenotyping of adult
  zebrafish behavior,'' {\em PLOS ONE}, vol.~6, no.~3, pp.~1--14, 2011.

\bibitem{cheng_3d_2016}
X.~E. Cheng, S.~H. Wang, and Y.~Q. Chen, ``{3D} tracking targets via kinematic
  model weighted particle filter,'' in {\em 2016 {IEEE} {International}
  {Conference} on {Multimedia} and {Expo} ({ICME})}, pp.~1--6, July 2016.

\bibitem{viscido_individual_2004}
S.~V. Viscido, J.~K. Parrish, and D.~Grünbaum, ``Individual behavior and
  emergent properties of fish schools: a comparison of observation and
  theory,'' {\em Marine Ecology Progress Series}, vol.~273, pp.~239--249, 2004.

\bibitem{straw_multi-camera_2010}
A.~D. Straw, K.~Branson, T.~R. Neumann, and M.~H. Dickinson, ``Multi-camera
  real-time three-dimensional tracking of multiple flying animals,'' {\em
  Journal of The Royal Society Interface}, vol.~8, pp.~395--409, July 2010.

\bibitem{muller_calibration_2014}
K.~Müller, J.~Schlemper, L.~Kuhnert, and K.~D. Kuhnert, ``Calibration and {3D}
  ground truth data generation with orthogonal camera-setup and refraction
  compensation for aquaria in real-time,'' in {\em 2014 {International}
  {Conference} on {Computer} {Vision} {Theory} and {Applications} ({VISAPP})},
  vol.~3, pp.~626--634, Jan. 2014.

\bibitem{stewart_novel_2015}
A.~M. Stewart, F.~Grieco, R.~A.~J. Tegelenbosch, E.~J. Kyzar, M.~Nguyen,
  A.~Kaluyeva, C.~Song, L.~P. J.~J. Noldus, and A.~V. Kalueff, ``A novel {3D}
  method of locomotor analysis in adult zebrafish: Implications for automated
  detection of {CNS} drug-evoked phenotypes,'' {\em Journal of Neuroscience
  Methods}, vol.~255, pp.~66--74, Nov. 2015.

\bibitem{qian_effective_2016}
Z.-M. Qian, S.~H. Wang, X.~E. Cheng, and Y.~Q. Chen, ``An effective and robust
  method for tracking multiple fish in video image based on fish head
  detection,'' {\em BMC Bioinformatics}, vol.~17, p.~251, June 2016.

\bibitem{Liu2019}
X.~Liu, Y.~Yue, M.~Shi, and Z.-M. Qian, ``{3-D} video tracking of multiple fish
  in a water tank,'' {\em {IEEE} Access}, vol.~7, pp.~145049--145059, 2019.

\bibitem{Wang_LSTM_2D_2016}
S.~H. {Wang}, X.~E. {Cheng}, and Y.~Q. {Chen}, ``Tracking undulatory body
  motion of multiple fish based on midline dynamics modeling,'' in {\em 2016
  IEEE International Conference on Multimedia and Expo (ICME)}, pp.~1--6, July
  2016.

\bibitem{Wang_LSTM_3D_2017}
S.~H. {Wang}, J.~{Zhao}, X.~{Liu}, Z.~{Qian}, Y.~{Liu}, and Y.~Q. {Chen},
  ``{3D} tracking swimming fish school with learned kinematic model using
  {LSTM} network,'' in {\em 2017 IEEE International Conference on Acoustics,
  Speech and Signal Processing (ICASSP)}, pp.~1068--1072, March 2017.

\bibitem{wang_robust_2016}
S.~H. Wang, J.~W. Zhao, and Y.~Q. Chen, ``Robust tracking of fish schools using
  {CNN} for head identification,'' {\em Multimedia Tools and Applications},
  pp.~1--19, 2016.

\bibitem{Pedersen_2018_CVPR_Workshops}
M.~Pedersen, S.~Hein~Bengtson, R.~Gade, N.~Madsen, and T.~B. Moeslund, ``Camera
  calibration for underwater {3D} reconstruction based on ray tracing using
  {Snell's} law,'' in {\em The IEEE Conference on Computer Vision and Pattern
  Recognition (CVPR) Workshops}, June 2018.

\bibitem{prewitt_analysis_1966}
J.~M.~S. Prewitt and M.~L. Mendelsohn, ``The analysis of cell images,'' {\em
  Annals of the New York Academy of Sciences}, vol.~128, no.~3, pp.~1035--1053,
  1966.

\bibitem{zhang1984fast}
T.~Zhang and C.~Y. Suen, ``A fast parallel algorithm for thinning digital
  patterns,'' {\em Communications of the ACM}, vol.~27, no.~3, pp.~236--239,
  1984.

\bibitem{ren2015faster}
S.~Ren, K.~He, R.~Girshick, and J.~Sun, ``Faster {R-CNN}: Towards real-time
  object detection with region proposal networks,'' in {\em Advances in Neural
  Information Processing Systems}, pp.~91--99, 2015.

\bibitem{kuhn_hungarian_1955}
H.~W. Kuhn, ``The {Hungarian} method for the assignment problem,'' {\em Naval
  Research Logistics Quarterly}, vol.~2, no.~1-2, pp.~83--97, 1955.

\bibitem{MTBF}
P.~{Carr} and R.~T. {Collins}, ``Assessing tracking performance in complex
  scenarios using mean time between failures,'' in {\em 2016 IEEE Winter
  Conference on Applications of Computer Vision (WACV)}, pp.~1--10, March 2016.

\bibitem{CLEARMOT}
K.~Bernardin and R.~Stiefelhagen, ``Evaluating multiple object tracking
  performance: The {CLEAR} {MOT} metrics,'' {\em EURASIP Journal on Image and
  Video Processing}, vol.~2008, p.~246309, May 2008.

\bibitem{MTML_Metrics}
{Bo Wu} and R.~{Nevatia}, ``Tracking of multiple, partially occluded humans
  based on static body part detection,'' in {\em 2006 IEEE Computer Society
  Conference on Computer Vision and Pattern Recognition (CVPR'06)}, vol.~1,
  pp.~951--958, June 2006.

\bibitem{Ristani_id_metrics}
E.~Ristani, F.~Solera, R.~Zou, R.~Cucchiara, and C.~Tomasi, ``Performance
  measures and a data set for multi-target, multi-camera tracking,'' in {\em
  Computer Vision -- ECCV 2016 Workshops} (G.~Hua and H.~J{\'e}gou, eds.),
  (Cham), pp.~17--35, Springer International Publishing, 2016.

\bibitem{Prewitt1966}
J.~M.~S. Prewitt and M.~L. Mendelsohn, ``The analysis of cell images*,'' {\em
  Annals of the New York Academy of Sciences}, vol.~128, no.~3, pp.~1035--1053,
  1966.

\bibitem{zhang_fast_1984}
T.~Y. Zhang and C.~Y. Suen, ``A {Fast} {Parallel} {Algorithm} for {Thinning}
  {Digital} {Patterns},'' {\em Commun. ACM}, vol.~27, pp.~236--239, Mar. 1984.

\bibitem{SAHOO}
P.~Sahoo, S.~Soltani, and A.~Wong, ``A survey of thresholding techniques,''
  {\em Computer Vision, Graphics, and Image Processing}, vol.~41, no.~2,
  pp.~233 -- 260, 1988.

\bibitem{FPN}
T.-Y. Lin, P.~Dollar, R.~Girshick, K.~He, B.~Hariharan, and S.~Belongie,
  ``Feature pyramid networks for object detection,'' in {\em The IEEE
  Conference on Computer Vision and Pattern Recognition (CVPR)}, July 2017.

\bibitem{IDSwitch_metric}
Y.~{Li}, C.~{Huang}, and R.~{Nevatia}, ``Learning to associate: {HybridBoosted}
  multi-target tracker for crowded scene,'' in {\em 2009 IEEE Conference on
  Computer Vision and Pattern Recognition}, pp.~2953--2960, June 2009.

\end{thebibliography}
}

\appendix

\title{3D-ZeF: A 3D Zebrafish Tracking Benchmark Dataset\\\large{Supplementary Material}}
\date{\vspace{-1.5ex}}

\author{Malte Pedersen\footnotemark[1]~, Joakim Bruslund Haurum\footnotemark[1]~, Stefan Hein Bengtson, Thomas B. Moeslund\\
Visual Analysis of People (VAP) Laboratory, Aalborg University, Denmark\\
{\tt\small mape@create.aau.dk, joha@create.aau.dk, shbe@create.aau.dk, tbm@create.aau.dk}}
\maketitle
\section{Content}
In these supplementary materials we provide: examples of the visual difference between the fish in the three splits, a more detailed explanation of the detectors, details on the benchmark pipeline parameters, a more detailed presentation of the relation between the proposed dataset complexity measures and tracking metrics, and the full set of tracking metrics for each step in the proposed benchmark pipeline.

\section{Fish Examples}
\begin{figure*}[!ht]
    \centering
    \begin{subfigure}[b]{0.3\textwidth}
        \centering
        \includegraphics[width=\textwidth]{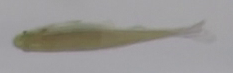}
        \caption{Fish from Trn5 top-view.}
        \label{fig:trn5top}
    \end{subfigure}
    \begin{subfigure}[b]{0.3\textwidth}
        \centering
        \includegraphics[width=\textwidth]{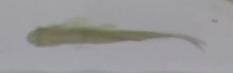}
        \caption{Fish from Val5 top-view.}
        \label{fig:vld5top}
    \end{subfigure}
    \begin{subfigure}[b]{0.3\textwidth}
        \centering
        \includegraphics[width=\textwidth]{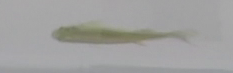}
        \caption{Fish from Tst10 top-view.}
        \label{fig:vld5top}
    \end{subfigure}
    \begin{subfigure}[b]{0.3\textwidth}
        \centering
        \includegraphics[width=\textwidth]{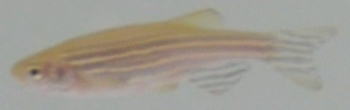}
        \caption{Fish from Trn5 front-view.}
        \label{fig:trn5front}
    \end{subfigure}
    \begin{subfigure}[b]{0.3\textwidth}
        \centering
        \includegraphics[width=\textwidth]{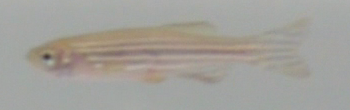}
        \caption{Fish from Val5 front-view.}
        \label{fig:vld5front}
    \end{subfigure}
    \begin{subfigure}[b]{0.3\textwidth}
        \centering
        \includegraphics[width=\textwidth]{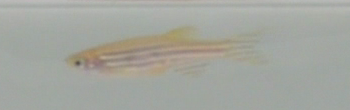}
        \caption{Fish from Tst10 front-view.}
        \label{fig:tst10front}
    \end{subfigure}
    \caption{Examples of zebrafish from the train, validation, and test split captured from the approximately same position in the water tank. The top- and front-view pairs show the same fish. Notice the variation in size of the zebrafish.}
    \label{fig:fish_examples}
\end{figure*}

The visual appearance of the fish varies both within and between the splits as unique groups of fish have been used for each split.
However, the two groups of fish used in the train and validation sets are from the same cohort, whereas the zebrafish in the test split are from a cohort of younger and smaller fish.
We present a fish from each of the three splits in \figref{fish_examples} captured at the approximately same position in the water tank.
The resolution of the zebrafish from the test split is significantly smaller compared to the fish in the training and validation splits, mainly due to the physical size of the fish.

\section{Naive Object Detection}
The Naive object detection algorithms presented briefly in the paper are described in more details in this section. 
The methods are inspired by Qian \etal \cite{qian_effective_2016} and their work on tracking zebrafish in 2D.

\subsection{Pre-Processing}
Initially the background image, without any fish, is estimated by taking the median of $N$ images sampled uniformly across the video. 
In the tests presented in the paper a set of 80 images were used to create the background image for each recording.
All further processing is restricted within a defined region of interest based on the boundaries of the water tank and the level of water, in order to limit the processing time.

 The video is downsampled by a factor of $2$, in order to further decrease the processing time.
 Background subtraction is applied by calculating the absolute difference image, $|im-bg|$, choosing the max value across all color channels, and normalizing the image into the range $[0;255]$.  
 The resulting image is filtered with a $5\times 5$ median filter to reduce noise.

\subsection{Top-View Detection}
The top-view is thresholded based on the assumption of the image histogram being bimodal, due to a near uniform bright background, and darker zebrafish. Therefore the intermodal approach of Prewitt \etal \cite{Prewitt1966} can be utilized. The threshold is set to the middle point between the two modes in the image histogram. If the histogram is not bimodal, the frame is filtered with a $3\times 3$ mean filter, until its histogram is bimodal.

The fish are detected by applying a skeletonization based approach. BLOBs are initially detected and filled, and the skeletonization approach of Zhang and Suen \cite{zhang_fast_1984} is applied.
The skeleton is analyzed in order to find the skeleton keypoints: head, tail, and junctions.
This is done by convolving the frame with the kernel in \equationref{intPtMat}.
All end points of the skeleton have a value of 116, 117, 118, or 131, while junctions have a value of 148, 149, 150, or 151. 
Values, like 132, that can represent both an end point or an arbitrary point on the skeleton are not considered. 
\begin{equation}\label{eq:intPtMat}
\begin{bmatrix}
1 & 1 & 1 & 1 & 1 \\
1 & 15 & 15 & 15 & 1 \\
1 & 15 & 100 & 15 & 1 \\
1 & 15 & 15 & 15 & 1 \\
1 & 1 & 1 & 1 & 1
\end{bmatrix}
\end{equation}

Each keypoint is assigned a weight, $w$,  consisting of the smallest eigenvalue of the covariance matrix of the BLOB coordinates extracted from a  $20\times 20$ window around the keypoint.
The smallest eigenvalue is used, as it is assumed the variance will be smallest along the width of the head.
This way the head keypoints will be assigned a larger $w$ than the tail keypoints, as the width of the zebrafish head is usually larger than the width of the zebrafish tail.
The $w$ of the junction points are reduced by a factor of $2.5$, as they are usually larger than endpoint keypoints. 

Keypoints too close to each other are removed by applying non-maximal suppression (NMS). Each keypoint is assigned a bounding box of size $w\times w$, and kept if the bounding box does not overlap with any other keypoint by more than NMS$_{\text{thresh}}$\% of the area of the keypoint in focus.
In case there is an overlap, only the keypoint with the largest $w$ is kept. 
NMS$_{\text{thresh}}$ was set to 50 in the conducted tests. 

Finally, all keypoints with $w<1$ are discarded.
Per skeleton, the two keypoints furthest apart are determined, and the keypoint with largest $w$ is kept.
Furthermore, half of the additional extra keypoints with the largest $w$ values are also kept in order to handle crossings and occlusions.
Every keypoint ideally represent a zebrafish head and they are all kept as individual detections.

\begin{table}[t!]
\centering
\begin{tabular}{lccccccc}
\hline
Parameter & $c$ & $\delta_T$ & $\delta_F$ & $\tau_k$ & $\alpha$ & $\tau_p$ & $\beta$ \\
Value & 95 & 15 & 0.5/15 & 10 & 10 & 25 & 0.02 \\ \hline
\end{tabular}
\caption{Pipeline parameters used for testing.}
\label{tab:parameters}
\end{table}

\subsection{Front-View Detection}
The front frame cannot be thresholded based on the assumption of a bimodal distribution of the image histogram, as the stripes of the zebrafish results in a non-uniform appearance.
Instead the frame is thresholded by finding the point maximizing the total entropy of the image \cite{SAHOO}.
The entropy is modeled as \textit{background} and \textit{foreground} entropy for each non-zero bin.

For bin $k$ the background entropy, $b_k$, is calculated by 
\begin{equation}
b_k = |\sum_{i = 0}^k \frac{h(i)}{h_c(i)} \log(\frac{h(i)}{h_c(i)})|,
\end{equation}
where $h$ is the normalized image histogram and $h_c$ is the cumulative histogram of $h$.
The foreground entropy, $w_k$, is calculated by
\begin{equation}\label{eq:whiteEntropy}
w_k = |\sum_{i = k+1}^{255} \frac{h(i)}{1-h_c(i)} \log(\frac{h(i)}{1-h_c(i)})|,
\end{equation}
resulting in the two entropy sets $B = \{b_0,b_1,..,b_{255}\}$ and $W = \{w_0,w_1,..,w_{255}\}$ which are combined into
\begin{equation}\label{eq:combinedEntropy}
E = B+W,
\end{equation}
and the threshold, $t$, is then finally determined by
\begin{equation}\label{eq:entropyArgmin}
t = \argmax_x E(x).
\end{equation}

The BLOBs in the thresholded image are found through simple Connected Component Analysis (CCA).
Only the $2N_{\text{fish}}$ largest BLOBs are kept, as long as the area of the BLOBs are larger than a predefined threshold.
This allows for potential reflections being detected alongside the true detection of the fish.
Otherwise, if only the $N_{\text{fish}}$ largest BLOBs are considered for further analysis, it is possible that reflections are considered at the expense of real fish.

As it is not known where the head of the fish is, two points are saved as proxies for the head, together with the center-of-mass.
If the width of the BLOB bounding box is larger than the height, the proxy points are $(\min(x),\mu(y))$ and $(\max(x),\mu(y))$, where x and y are the coordinates of the BLOB pixels, and $\mu$ is the arithmetic mean function.
Otherwise, the proxy points are $(\mu(x),\min(y))$ and $(\mu(x),\max(y))$.
The center-of-mass is further used for constructing 2D tracklets, as it is the most stable of the three proxy points.\\

\section{FRCNN-H Object Detection}
Two Faster R-CNN networks were trained to detect the zebrafish heads in each view, respectively.
The official PyTorch implementation of Faster R-CNN with a ResNet50 based Feature Pyramid Network \cite{FPN} backbone was utilized in both cases.
The model was pretrained on the COCO train2017 dataset.
The network was fine-tuned for 30 epochs, using stochastic gradient descent with momentum.
A learning rate of 0.005, momentum of 0.9, weight decay of 0.0005, and batch size of 8 was used.
The learning rate was ``warmed up'' during the first epoch, linearly interpolating the learning rate from 0.001 to 0.005. 
Each model was trained on an RTX 2080TI.

\begin{table}[t!]
\centering
\begin{tabular}{llccc}
\hline
 & Property & Mean & Std. Dev. & Median  \\ \hline
\multirow{2}{*}{\STAB{\rotatebox[origin=c]{90}{Trn}}} 
 & Reproj. Error $[px]$ & 8.03 & 5.26 & 7.59\\
 & Speed $[cm / s^2]$ & 2.13 & 2.32 & 1.54\\ \hline
\multirow{2}{*}{\STAB{\rotatebox[origin=c]{90}{Val}}} 
 & Reproj. Error $[px]$ & 6.22 & 4.50 & 5.43\\
 & Speed $[cm / s^2]$  & 2.02 & 2.37 & 1.41\\ \hline
\multirow{2}{*}{\STAB{\rotatebox[origin=c]{90}{Tst}}} 
 & Reproj. Error $[px]$ & 4.36 & 3.27 & 3.69\\
 & Speed $[cm / s^2]$ & 2.11 & 1.93 & 1.58\\ \hline
\end{tabular}
\caption{Statistics of the reprojection error and speed of the zebrafish, for each split, based on the ground truth annotations.}
\label{tab:splitStats}
\end{table}

\section{Pipeline Parameters}
The full system pipeline has a set of parameters, which needs to be manually set.
The parameters were chosen based on empirical investigation on the training data and they are shown in \tableref{parameters}. 
The front-view distance threshold, $\delta_F$, has two different values, depending on the method applied.
For the Naive method it is set to $0.5$, as the distance is measured in standard deviations, whereas for the FRCNN-H method it is set to $15$, where the distance is measured in pixels. 

Furthermore, during the association of 2D tracklets between views, a set of exponential distributions are utilized. These distributions are parameterized using the mean of the reprojection error of the ground truth training annotations and the average speed of the zebrafish in the training split. This builds on the assumption that these parameters generalize across the different splits. As shown in \tableref{splitStats}, the average speed of the zebrafish  for each split are within $0.1 \frac{cm}{s^2}$, which can be seen as a negligible difference, whereas the mean reprojection error is $2-4$ pixels larger in the training and validation splits than in the testing split. The cause of this difference is hypothesized to be the larger amount and duration of occlusions in the training and validation splits, which can introduce some uncertainty during the annotation phase. The training split has the largest reprojection error, which means the utilized exponential distribution penalize larger reprojection errors less harshly. However, as the mean training reprojection error is still low, it is still deemed fitting for the task.

\begin{figure}[!t]
    \centering
    \includegraphics[width=0.45\textwidth]{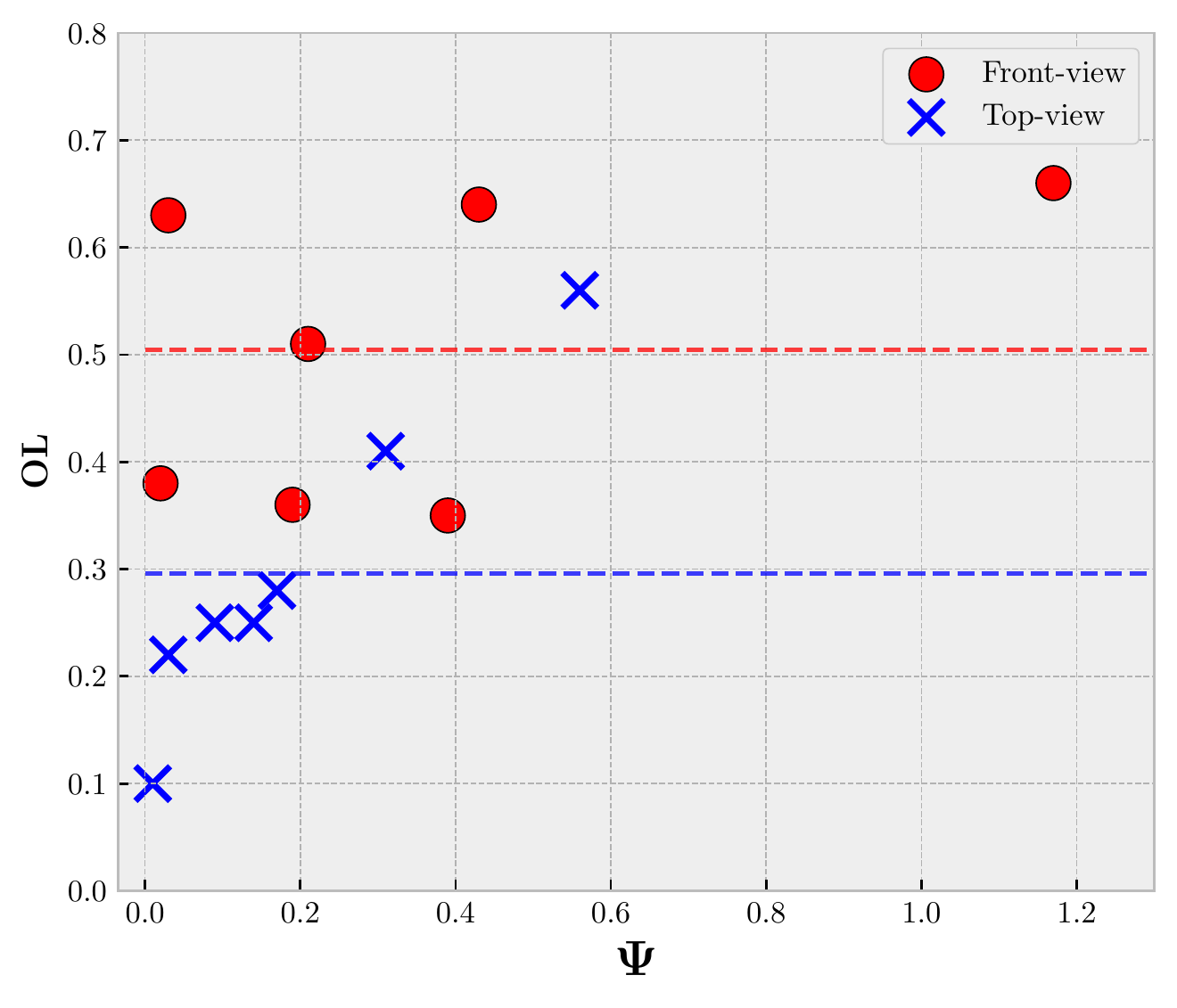}
    \caption{$\Psi$ is the proposed complexity measure based on the front- and top-view recordings. OL is the average length of occlusion events and the dashed lines show the mean OL for the two views. Tst1 is not included.}
    \label{fig:occ_ol}
\end{figure}

\section{Tracking Metrics and Results}
When evaluating tracking tasks, a wide suite of metrics are often used, each expressing different properties of the task.
Commonly, the MOTA metric is used as a representative metric for the overall performance of the tracker, as it incorporates false positives and negatives, as well as the amount of identity swaps.
However, as shown by Carr and Collins \cite{MTBF} the MOTA metric is not guaranteed  to be robust at all times.
We evaluate the generated tracks using the collection of metrics from the MOT challenge, consisting of the CLEAR MOT \cite{CLEARMOT}, mostly tracked/lost metric \cite{MTML_Metrics}, and identity based metrics of Ristani \etal \cite{Ristani_id_metrics}, as well as using the MTBF metric proposed by Carr and Collins \cite{MTBF}. 
A description of each of the used metrics are presented in \tableref{trackmetrics}.
Each step of the pipeline has been evaluated according to the previous mentioned metrics. 
Specifically the following steps have been evaluated:
\begin{itemize}
    \item 2D tracklets after 2D Tracklet Construction (for top-view see \tableref{trackMeasuresTop} and for front-view see \tableref{trackMeasuresFront})
\item 2D tracklets after 2D Tracklet Association Between Views (for top-view see \tableref{trackMeasuresTop3D} and for front-view see \tableref{trackMeasuresFront3D})
\item 3D tracklets after 2D Tracklet Association Between Views (see \tableref{trackMeasures3DTracklets})
\item 3D tracks after 3D Tracklet Association (see \tableref{trackMeasures3DTracks})
\end{itemize}
For the 2D tracklets, a distance threshold of 20 px is used.

\begin{figure}[!t]
    \centering
    \includegraphics[width=0.45\textwidth]{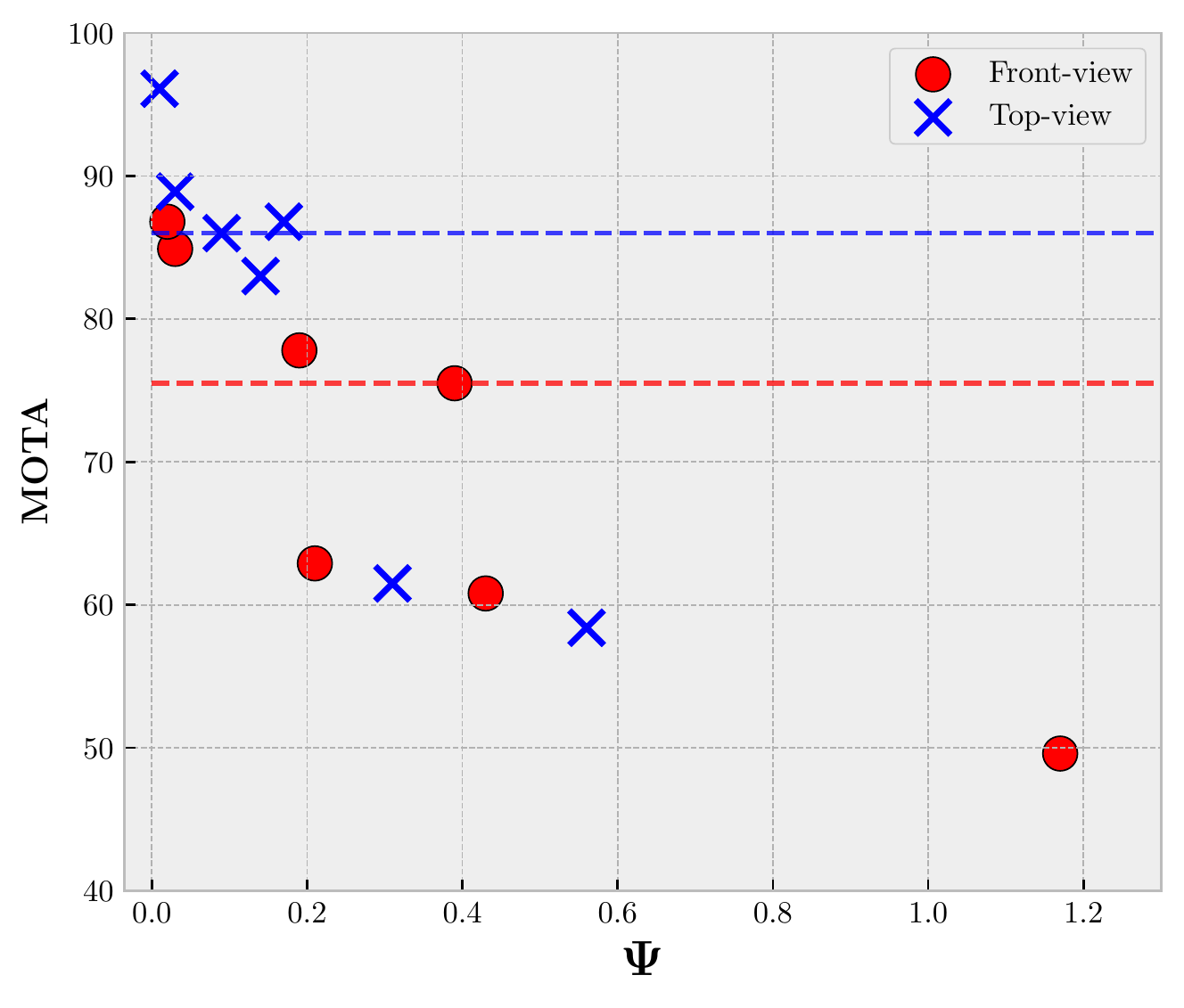}
    \caption{$\Psi$ is the proposed complexity measure based on the front- and top-view recordings. MOTA is based on the oracle tracker and the dashed lines show the mean MOTA for the two views. Tst1 is not included.}
    \label{fig:occ_mota}
\end{figure}

\section{Occlusions}
The two views are not equal when it comes to occlusions.
Due to the social behavior of zebrafish they tend to swim in the same horizontal layer of the water column.
This is indicated by the graph in \figref{occ_ol}, where the occlusion lengths are displayed with respect to the proposed complexity measure, $\Psi$. 
It should be noted that $\Psi$ is calculated per view and not as a mean between the two, as in the main article.
Data points are presented for all the recordings from both views, except Tst1 as it only contains a single fish and therefore has a complexity measure of zero. 
The dashed lines illustrate the mean occlusion lengths and shows that there is a significant difference between the two views.
In general, the occlusion events seems longer in the front-view recordings.

The MOTA of the hypothetical oracle tracker is plotted against the proposed complexity measure in \figref{occ_mota}.
MOTA is calculated for the top- and front-view 2D tracklets as presented in \tableref{trackMeasuresTop} and \tableref{trackMeasuresFront}, respectively.
The tracking performance of the oracle tracker correlates well with the complexity in both views.
Notice the significant difference in performance of the tracker in the two views; the mean scores are MOTA$_{\text{top}}=86.0\%$ and MOTA$_{\text{front}}=75.5\%$ as illustrated by the two dashed lines.
\begin{table*}[]
\centering
\begin{tabular}{llll}
\hline
Metric &  Better & Range & Description \\
\hline
MOTA        & Higher    & $]-\infty ,100]$      & Overall tracking accuracy based on ID-swaps, and false negatives/positives \cite{CLEARMOT}\\
MOTP        & Lower     & $[0 ,\infty[$         & Distance between predicted position and ground truth \cite{CLEARMOT}\\
Prc.        & Higher    & $[0 ,100]$            & Precision: The percentage of correctly predicted positions \\
Rcll.       & Higher    & $[0 ,100]$            & Recall: The percentage of ground truth positions detected \\
ID-Prc.     & Higher    & $[0 ,100]$            & ID-Precision: The percentage of correctly identified predicted positions \cite{Ristani_id_metrics}\\
ID-Rcll.    & Higher    & $[0 ,100]$            & ID-Recall:  The percentage of correctly identified ground truth positions \cite{Ristani_id_metrics}\\
ID-F1       & Higher    & $[0 ,100]$            & F1-score of the predicted identities \cite{Ristani_id_metrics}\\
FP          & Lower     & $[0 ,\infty[$         & Amount of incorrectly predicted positions\\
FN          & Lower     & $[0 ,\infty[$         & Amount of ground truth positions missed\\
MT          & Higher    & $[0 ,100]$            & Amount of ground truth tracks which are 80\% or more correctly tracked \cite{MTML_Metrics} \\
ML          & Lower     & $[0 ,100]$            & Amount of ground truth tracks which are 20\% or less correctly tracked \cite{MTML_Metrics} \\
ID Sw.      & Lower     & $[0 ,\infty[$         & Total amount of identity switches \cite{IDSwitch_metric} \\
Frag.       & Lower     & $[0 ,\infty[$         & Total amount of track fragmentations \\
MTBF$_{s}$  & Higher    & $[0 ,\infty[$         & Mean time between ID-switches or false negatives/positives \cite{MTBF}\\
MTBF$_{m}$  & Higher    & $[0 ,\infty[$         & Monotonic version of $\textrm{MTBF}_{s}$ \cite{MTBF}\\ \hline
\end{tabular}
\caption{Description of the full collection of utilized tracking metrics.}
\label{tab:trackmetrics}
\end{table*}

\begin{table*}[t!]
\centering
\resizebox{\linewidth}{!}{%
\begin{tabular}{clccccccccccccccc}
\hline
 & Method & MOTA $\uparrow$ & MOTP $\downarrow$ & Prc. $\uparrow$ & Rcll. $\uparrow$ & ID-Rcll. $\uparrow$ & ID-Prc. $\uparrow$ & ID-F1 $\uparrow$ & FP $\downarrow$ & FN $\downarrow$ & MT $\uparrow$ & ML $\downarrow$ & ID Sw. $\downarrow$ & Frag. $\downarrow$ & $\textrm{MTBF}_{s}$ $\uparrow$ & $\textrm{MTBF}_{m}$ $\uparrow$ \\ \hline
\multirow{3}{*}{\STAB{\rotatebox[origin=c]{90}{Trn2}}} 
 & Naive        & 21.9\% & 3.309 & 57.2 & 91.0 & 10.7 & 6.7 & 8.2 & 9798 & 1301 & 2 & 0 & 136 & 428 & 26.316 & 14.109 \\
 & FRCNN-H      & 91.3\% & 2.128 & 98.8 & 92.8 & 12.4 & 13.2 & 12.8 & 156 & 1030 & 2 & 0 & 72 & 140 & 87.829 & 45.408 \\ 
 & Oracle       & 61.5\% & 0.000 & 100.0 & 63.0 & 6.7 & 10.6 & 8.2 & 0 & 5314 & 0 & 0 & 216 & 216 & 41.587 & 20.699 \\ \hline
 \multirow{3}{*}{\STAB{\rotatebox[origin=c]{90}{Trn5}}} 
 & Naive        & 38.2\% & 3.196 & 64.1 & 90.1 & 31.5 & 22.4 & 26.2 & 2274 & 447 & 5 & 0 & 58 & 124 & 27.020 & 14.792 \\
 & FRCNN-H      & 90.2\% & 2.038 & 99.1 & 91.6 & 31.9 & 34.5 & 33.2 & 39 & 378 & 5 & 0 & 24 & 50 & 73.607 & 38.523 \\
 & Oracle       & 58.4\% & 0.000 & 100.0 & 59.6 & 19.1 & 32.0 & 23.9 & 0 & 1819 & 0 & 0 & 52 & 52 & 47.035 & 23.726 \\ \hline
 \multirow{3}{*}{\STAB{\rotatebox[origin=c]{90}{Val2}}} 
 & Naive        & 59.8\% & 3.050 & 72.6 & 96.7 & 57.1 & 42.8 & 48.9 & 1315 & 118 & 2 & 0 & 13 & 44 & 68.275 & 36.653 \\
 & FRCNN-H      & 96.8\% & 2.165 & 99.2 & 97.9 & 75.9 & 76.8 & 76.3 & 30 & 76 & 2 & 0 & 10 & 17 & 167.810 & 92.737 \\
 & Oracle       & 88.9\% & 0.000 & 100.0 & 89.7 & 25.8 & 28.8 & 27.2 & 0 & 372 & 2 & 0 & 28 & 28 & 107.600 & 55.655 \\ \hline
 \multirow{3}{*}{\STAB{\rotatebox[origin=c]{90}{Val5}}} 
 & Naive        & 89.4\% & 3.552 & 93.1 & 97.0 & 51.7 & 49.6 & 50.6 & 330 & 135 & 5 & 0 & 19 & 36 & 92.083 & 52.619 \\
 & FRCNN-H      & 96.4\% & 2.782 & 99.5 & 97.2 & 48.3 & 49.4 & 48.8 & 22 & 128 & 5 & 0 & 15 & 23 & 147.567 & 83.528 \\
 & Oracle       & 86.0\% & 0.000 & 100.0 & 86.9 & 37.2 & 42.8 & 39.8 & 0 & 598 & 4 & 0 & 40 & 40 & 87.933 & 46.553 \\ \hline
 \multirow{3}{*}{\STAB{\rotatebox[origin=c]{90}{Tst1}}} 
 & Naive        &  83.1\% & 4.095 & 85.6 & 100.0 & 100.0 & 85.6 & 92.2 & 152 & 0 & 1 & 0 & 0 & 0 & 900.000 & 900.000 \\
 & FRCNN-H      &  85.1\% & 2.513 & 100.0 & 85.7 & 49.7 & 58.0 & 53.5 & 0 & 129 & 1 & 0 & 5 & 34 & 22.029 & 11.174 \\
 & Oracle       & 100.0\% & 0.000 & 100.0 & 100.0 & 100.0 & 100.0 & 100.0 & 0 & 0 & 1 & 0 & 0 & 0 & 900.000 & 900.000 \\ \hline
 \multirow{3}{*}{\STAB{\rotatebox[origin=c]{90}{Tst2}}} 
 & Naive        & 98.6\% & 3.771 & 99.6 & 99.1 & 99.1 & 99.6 & 99.3 & 8 & 17 & 2 & 0 & 0 & 10 & 148.583 & 81.045 \\
 & FRCNN-H      & 72.6\% & 2.443 & 100.0 & 74.1 & 39.0 & 52.7 & 44.8 & 0 & 467 & 1 & 0 & 26 & 93 & 14.032 & 7.053 \\
 & Oracle       & 96.1\% & 0.000 & 100.0 & 96.7 & 32.6 & 33.7 & 33.1 & 0 & 60 & 2 & 0 & 10 & 10 & 145.000 & 79.091 \\ \hline
 \multirow{3}{*}{\STAB{\rotatebox[origin=c]{90}{Tst5}}} 
 & Naive        & 82.8\% & 3.801 & 89.3 & 94.7 & 49.7 & 46.8 & 48.2 & 512 & 237 & 5 & 0 & 24 & 82 & 45.839 & 24.222 \\
 & FRCNN-H      & 70.5\% & 2.685 & 99.9 & 71.8 & 25.6 & 35.6 & 29.7 & 4 & 1271 & 0 & 0 & 52 & 145 & 21.384 & 10.836 \\
 & Oracle       & 83.0\% & 0.000 & 100.0 & 84.0 & 44.7 & 53.2 & 48.6 & 0 & 721 & 3 & 0 & 44 & 44 & 77.122 & 39.779 \\ \hline
 \multirow{3}{*}{\STAB{\rotatebox[origin=c]{90}{Tst10}}} 
 & Naive        & 83.4\% & 3.752 & 89.3 & 95.4 & 59.0 & 55.2 & 57.0 & 1026 & 418 & 10 & 0 & 49 & 138 & 52.329 & 28.323 \\
 & FRCNN-H      & 70.8\% & 2.803 & 99.8 & 72.4 & 26.4 & 36.4 & 30.6 & 13 & 2483 & 4 & 0 & 130 & 417 & 14.879 & 7.543 \\
 & Oracle       & 86.8\% & 0.000 & 100.0 & 87.5 & 42.1 & 48.1 & 44.9 & 0 & 1128 & 10 & 0 & 64 & 64 & 106.378 & 56.229 \\ \hline
\end{tabular}%
}
\caption{Full metric evaluation of the 2D tracking in the top-view, on all sequences.}
\label{tab:trackMeasuresTop}
\end{table*}

\begin{table*}[t!]
\centering
\resizebox{\linewidth}{!}{%
\begin{tabular}{clccccccccccccccc}
\hline
 & Method & MOTA $\uparrow$ & MOTP $\downarrow$ & Prc. $\uparrow$ & Rcll. $\uparrow$ & ID-Rcll. $\uparrow$ & ID-Prc. $\uparrow$ & ID-F1 $\uparrow$ & FP $\downarrow$ & FN $\downarrow$ & MT $\uparrow$ & ML $\downarrow$ & ID Sw. $\downarrow$ & Frag. $\downarrow$ & $\textrm{MTBF}_{s}$ $\uparrow$ & $\textrm{MTBF}_{m}$ $\uparrow$ \\ \hline
\multirow{3}{*}{\STAB{\rotatebox[origin=c]{90}{Trn2}}} 
 & Naive        & -69.5\% & 12.666 & 7.6 & 6.2 & 0.6 & 0.7 & 0.6 & 10756 & 13490 & 0 & 2 & 134 & 159 & 4.709 & 2.528 \\
 & FRCNN-H      &  92.8\% & 3.521 & 99.8 & 95.0 & 6.4 & 6.8 & 6.6 & 26 & 723 & 2 & 0 & 288 & 126 & 43.218 & 30.898 \\
 & Oracle       &  62.9\% & 0.000 & 100.0 & 64.1 & 2.4 & 3.8 & 3.0 & 0 & 5168 & 0 & 0 & 170 & 170 & 53.558 & 26.936 \\ \hline
\multirow{3}{*}{\STAB{\rotatebox[origin=c]{90}{Trn5}}} 
 & Naive        & -73.5\% & 12.768 & 5.2 & 4.2 & 3.1 & 3.8 & 3.4 & 3478 & 4310 & 0 & 5 & 19 & 22 & 5.758 & 2.923 \\
 & FRCNN-H      &  90.7\% & 3.634 & 99.9 & 92.4 & 23.2 & 25.1 & 24.1 & 6 & 341 & 5 & 0 & 72 & 47 & 45.207 & 29.496 \\
 & Oracle       &  60.8\% & 0.000 & 100.0 & 61.7 & 20.1 & 32.6 & 24.9 & 0 & 1724 & 1 & 0 & 41 & 41 & 60.348 & 29.849 \\ \hline
\multirow{3}{*}{\STAB{\rotatebox[origin=c]{90}{Val2}}} 
 & Naive        & -93.1\% & 12.918 & 3.7 & 3.6 & 0.8 & 0.8 & 0.8 & 3444 & 3469 & 0 & 2 & 37 & 28 & 3.119 & 1.795 \\
 & FRCNN-H      &  93.6\% & 4.113 & 100.0 & 97.4 & 12.9 & 13.3 & 13.1 & 0 & 95 & 2 & 0 & 136 & 24 & 25.036 & 21.372 \\ 
 & Oracle       &  84.9\% & 0.000 & 100.0 & 85.3 & 16.3 & 19.1 & 17.6 & 0 & 530 & 2 & 0 & 14 & 14 & 191.875 & 102.333 \\ \hline
\multirow{3}{*}{\STAB{\rotatebox[origin=c]{90}{Val5}}} 
 & Naive        & -68.8\% & 12.419 & 10.0 & 8.5 & 2.9 & 3.4 & 3.1 & 3480 & 4162 & 0 & 5 & 37 & 44 & 6.690 & 3.464 \\
 & FRCNN-H      &  79.0\% & 4.224 & 99.6 & 81.8 & 16.2 & 19.7 & 17.8 & 14 & 826 & 3 & 0 & 117 & 72 & 26.791 & 17.402 \\ 
 & Oracle       &  49.6\% & 0.000 & 100.0 & 50.8 & 16.5 & 32.4 & 21.8 & 0 & 2240 & 0 & 0 & 54 & 54 & 39.153 & 19.914 \\ \hline
\multirow{3}{*}{\STAB{\rotatebox[origin=c]{90}{Tst1}}} 
 & Naive        & -44.8\% & 14.682 & 27.5 & 26.2 & 5.2 & 5.5 & 5.4 & 621 & 664 & 0 & 0 & 18 & 28 & 6.378 & 3.522 \\
 & FRCNN-H      &  41.4\% & 7.031 & 99.0 & 44.4 & 15.0 & 33.4 & 20.7 & 4 & 500 & 0 & 0 & 23 & 17 & 13.333 & 8.163 \\ 
 & Oracle       & 100.0\% & 0.000 & 100.0 & 100.0 & 100.0 & 100.0 & 100.0 & 0 & 0 & 1 & 0 & 0 & 0 & 900.000 & 900.000 \\ \hline
\multirow{3}{*}{\STAB{\rotatebox[origin=c]{90}{Tst2}}} 
 & Naive        & -38.3\% & 13.201 & 30.0 & 27.0 & 5.1 & 5.6 & 5.3 & 1135 & 1314 & 0 & 1 & 41 & 41 & 7.967 & 4.673 \\
 & FRCNN-H      &  16.4\% & 7.828 & 100.0 & 18.7 & 6.6 & 35.4 & 11.1 & 0 & 1464 & 0 & 1 & 40 & 57 & 5.695 & 2.824 \\ 
 & Oracle       &  86.8\% & 0.000 & 100.0 & 87.3 & 45.8 & 52.4 & 48.9 & 0 & 228 & 2 & 0 & 10 & 10 & 131.000 & 71.455 \\ \hline
\multirow{3}{*}{\STAB{\rotatebox[origin=c]{90}{Tst5}}} 
 & Naive        & -39.2\% & 12.703 & 28.5 & 24.3 & 4.6 & 5.4 & 4.9 & 2734 & 3408 & 0 & 3 & 124 & 138 & 5.717 & 3.250 \\
 & FRCNN-H      &  53.5\% & 5.601 & 98.7 & 56.4 & 15.4 & 27.0 & 19.7 & 33 & 1962 & 0 & 0 & 99 & 147 & 15.962 & 8.161 \\ 
 & Oracle       &  77.8\% & 0.000 & 100.0 & 78.7 & 32.6 & 41.4 & 36.5 & 0 & 957 & 2 & 0 & 42 & 42 & 75.383 & 38.934 \\ \hline
\multirow{3}{*}{\STAB{\rotatebox[origin=c]{90}{Tst10}}} 
 & Naive        & -54.1\% & 13.667 & 19.6 & 16.9 & 3.8 & 4.4 & 4.1 & 6219 & 7482 & 0 & 5 & 171 & 195 & 5.816 & 3.209 \\
 & FRCNN-H      &  47.4\% & 5.604 & 99.4 & 50.3 & 13.0 & 25.6 & 17.2 & 27 & 4474 & 0 & 0 & 229 & 277 & 14.368 & 7.518 \\ 
 & Oracle       &  75.5\% & 0.000 & 100.0 & 76.5 & 25.7 & 33.6 & 29.2 & 0 & 2115 & 4 & 0 & 94 & 94 & 66.202 & 33.261 \\ \hline
\end{tabular}%
}
\caption{Full metric evaluation of the 2D tracking in the front-view, on all sequences.}
\label{tab:trackMeasuresFront}
\end{table*}

\begin{table*}[t!]
\centering
\resizebox{\linewidth}{!}{%
\begin{tabular}{clccccccccccccccc}
\hline
 & Method & MOTA $\uparrow$ & MOTP $\downarrow$ & Prc. $\uparrow$ & Rcll. $\uparrow$ & ID-Rcll. $\uparrow$ & ID-Prc. $\uparrow$ & ID-F1 $\uparrow$ & FP $\downarrow$ & FN $\downarrow$ & MT $\uparrow$ & ML $\downarrow$ & ID Sw. $\downarrow$ & Frag. $\downarrow$ & $\textrm{MTBF}_{s}$ $\uparrow$ & $\textrm{MTBF}_{m}$ $\uparrow$ \\ \hline
\multirow{3}{*}{\STAB{\rotatebox[origin=c]{90}{Trn2}}} 
 & Naive        & 82.6\% & 3.247 & 96.2 & 86.5 & 13.2 & 14.6 & 13.9 & 497 & 1939 & 2 & 0 & 73 & 403 & 28.732 & 14.846 \\
 & FRCNN-H      & 90.9\% & 2.113 & 100.0 & 91.1 & 22.1 & 24.2 & 23.1 & 1 & 1273 & 2 & 0 & 41 & 144 & 85.667 & 43.836 \\
 & Oracle       & 44.8\% & 0.000 & 100.0 & 46.2 & 2.4 & 5.3 & 3.3 & 0 & 7742 & 0 & 0 & 202 & 202 & 32.539 & 16.190 \\ \hline
\multirow{3}{*}{\STAB{\rotatebox[origin=c]{90}{Trn5}}} 
 & Naive        & 79.3\% & 3.151 & 95.9 & 83.4 & 37.5 & 43.2 & 40.1 & 159 & 748 & 3 & 0 & 23 & 119 & 29.543 & 15.129 \\
 & FRCNN-H      & 90.2\% & 2.028 & 100.0 & 90.5 & 45.6 & 50.4 & 47.8 & 0 & 427 & 5 & 0 & 13 & 44 & 81.460 & 42.874 \\
 & Oracle       & 36.6\% & 0.000 & 100.0 & 37.8 & 10.9 & 28.9 & 15.9 & 0 & 2799 & 0 & 0 & 53 & 53 & 29.328 & 14.175 \\ \hline
\multirow{3}{*}{\STAB{\rotatebox[origin=c]{90}{Val2}}} 
 & Naive        & 89.0\% & 3.047 & 99.4 & 89.8 & 63.3 & 70.1 & 66.6 & 20 & 369 & 2 & 0 & 8 & 52 & 57.696 & 29.642 \\
 & FRCNN-H      & 96.1\% & 2.163 & 100.0 & 96.3 & 78.4 & 81.4 & 79.8 & 0 & 133 & 2 & 0 & 7 & 20 & 157.591 & 82.548 \\
 & Oracle       & 73.9\% & 0.000 & 100.0 & 74.9 & 12.7 & 17.0 & 14.5 & 0 & 902 & 0 & 0 & 36 & 36 & 71.000 & 35.500 \\\hline
\multirow{3}{*}{\STAB{\rotatebox[origin=c]{90}{Val5}}} 
 & Naive        & 92.1\% & 3.524 & 99.8 & 92.5 & 53.3 & 57.4 & 55.3 & 9 & 341 & 4 & 0 & 9 & 57 & 67.968 & 34.826 \\
 & FRCNN-H      & 90.9\% & 2.765 & 100.0 & 91.1 & 51.8 & 56.8 & 54.2 & 0 & 404 & 4 & 0 & 10 & 24 & 143.138 & 76.870 \\
 & Oracle       & 42.2\% & 0.000 & 100.0 & 43.6 & 13.2 & 30.3 & 18.4 & 0 & 2567 & 0 & 0 & 64 & 64 & 28.812 & 14.406 \\ \hline
\multirow{3}{*}{\STAB{\rotatebox[origin=c]{90}{Tst1}}} 
 & Naive        &  99.6\% & 4.102 & 100.0 & 99.6 & 99.6 & 100.0 & 99.8 & 0 & 4 & 1 & 0 & 0 & 2 & 298.667 & 179.200 \\
 & FRCNN-H      &  69.8\% & 2.477 & 100.0 & 69.8 & 69.8 & 100.0 & 82.2 & 0 & 272 & 0 & 0 & 0 & 31 & 19.625 & 9.812 \\
 & Oracle       & 100.0\% & 0.000 & 100.0 & 100.0 & 100.0 & 100.0 & 100.0 & 0 & 0 & 1 & 0 & 0 & 0 & 900.000 & 900.000 \\ \hline
\multirow{3}{*}{\STAB{\rotatebox[origin=c]{90}{Tst2}}} 
 & Naive        & 98.1\% & 3.782 & 99.9 & 98.1 & 98.1 & 99.9 & 99.0 & 1 & 34 & 2 & 0 & 0 & 18 & 88.300 & 46.474 \\
 & FRCNN-H      & 40.8\% & 2.514 & 100.0 & 40.9 & 32.8 & 80.1 & 46.5 & 0 & 1063 & 0 & 1 & 2 & 35 & 19.919 & 9.827 \\
 & Oracle       & 82.2\% & 0.000 & 100.0 & 83.2 & 24.3 & 29.2 & 26.6 & 0 & 302 & 2 & 0 & 18 & 18 & 74.900 & 37.450 \\ \hline
\multirow{3}{*}{\STAB{\rotatebox[origin=c]{90}{Tst5}}} 
 & Naive        & 86.1\% & 3.737 & 98.3 & 87.9 & 52.8 & 59.1 & 55.8 & 67 & 546 & 4 & 0 & 13 & 69 & 51.351 & 26.716 \\
 & FRCNN-H      & 65.8\% & 2.666 & 100.0 & 66.2 & 34.0 & 51.3 & 40.9 & 0 & 1519 & 0 & 0 & 21 & 110 & 25.698 & 13.075 \\
 & Oracle       & 65.7\% & 0.000 & 100.0 & 66.8 & 26.8 & 40.1 & 32.2 & 0 & 1492 & 1 & 0 & 50 & 50 & 54.691 & 26.857 \\ \hline
\multirow{3}{*}{\STAB{\rotatebox[origin=c]{90}{Tst10}}} 
 & Naive        & 87.0\% & 3.757 & 98.4 & 88.6 & 68.1 & 75.7 & 71.7 & 126 & 1023 & 9 & 0 & 23 & 151 & 48.054 & 24.697 \\
 & FRCNN-H      & 62.4\% & 2.797 & 100.0 & 62.9 & 31.4 & 49.9 & 38.6 & 2 & 3341 & 2 & 0 & 39 & 293 & 18.677 & 9.323 \\
 & Oracle       & 64.7\% & 0.000 & 100.0 & 66.0 & 20.0 & 30.3 & 24.1 & 0 & 3059 & 1 & 0 & 119 & 119 & 46.054 & 22.676 \\ \hline
\end{tabular}%
}
\caption{Full metric evaluation of the 2D tracking in the top-view, after 3D tracklet association, on all sequences.}
\label{tab:trackMeasuresTop3D}
\end{table*}

\begin{table*}[t!]
\centering
\resizebox{\linewidth}{!}{%
\begin{tabular}{clccccccccccccccc}
\hline
 & Method & MOTA $\uparrow$ & MOTP $\downarrow$ & Prc. $\uparrow$ & Rcll. $\uparrow$ & ID-Rcll. $\uparrow$ & ID-Prc. $\uparrow$ & ID-F1 $\uparrow$ & FP $\downarrow$ & FN $\downarrow$ & MT $\uparrow$ & ML $\downarrow$ & ID Sw. $\downarrow$ & Frag. $\downarrow$ & $\textrm{MTBF}_{s}$ $\uparrow$ & $\textrm{MTBF}_{m}$ $\uparrow$ \\ \hline
\multirow{3}{*}{\STAB{\rotatebox[origin=c]{90}{Trn2}}} 
 & Naive        & -5.5\% & 14.688 & 46.6 & 33.4 & 6.6 & 9.2 & 7.7 & 5513 & 9577 & 0 & 0 & 82 & 772 & 6.174 & 3.091 \\
 & FRCNN-H      & 85.3\% & 3.482 & 99.9 & 85.6 & 19.5 & 22.8 & 21.0 & 10 & 2074 & 2 & 0 & 37 & 141 & 82.040 & 42.289 \\
 & Oracle       & 44.8\% & 0.000 & 100.0 & 46.2 & 2.4 & 5.3 & 3.3 & 0 & 7742 & 0 & 0 & 202 & 202 & 32.539 & 16.190 \\ \hline
\multirow{3}{*}{\STAB{\rotatebox[origin=c]{90}{Trn5}}} 
 & Naive        &  2.2\% & 13.648 & 51.9 & 37.8 & 19.1 & 26.2 & 22.1 & 1577 & 2800 & 0 & 2 & 22 & 168 & 9.770 & 4.843 \\
 & FRCNN-H      & 81.5\% & 3.540 & 99.9 & 82.0 & 41.4 & 50.5 & 45.5 & 3 & 812 & 3 & 0 & 16 & 47 & 68.296 & 35.124 \\
 & Oracle       & 36.6\% & 0.000 & 100.0 & 37.8 & 10.9 & 28.9 & 15.9 & 0 & 2799 & 0 & 0 & 53 & 53 & 29.328 & 14.058 \\ \hline
\multirow{3}{*}{\STAB{\rotatebox[origin=c]{90}{Val2}}} 
 & Naive        & -7.9\% & 14.061 & 45.5 & 39.0 & 30.7 & 35.8 & 33.0 & 1678 & 2197 & 0 & 0 & 11 & 165 & 8.401 & 4.201 \\
 & FRCNN-H      & 82.1\% & 4.095 & 100.0 & 82.3 & 63.6 & 77.2 & 69.7 & 0 & 638 & 1 & 0 & 7 & 36 & 75.949 & 39.493 \\
 & Oracle       & 73.9\% & 0.000 & 100.0 & 74.9 & 12.7 & 17.0 & 14.5 & 0 & 902 & 0 & 0 & 36 & 36 & 71.000 & 35.500 \\ \hline
\multirow{3}{*}{\STAB{\rotatebox[origin=c]{90}{Val5}}} 
 & Naive        & -6.9\% & 11.925 & 43.9 & 24.0 & 13.9 & 25.5 & 18.0 & 1391 & 3460 & 0 & 2 & 15 & 108 & 9.646 & 4.781 \\
 & FRCNN-H      & 69.3\% & 4.068 & 99.9 & 69.7 & 35.0 & 50.1 & 41.2 & 4 & 1379 & 0 & 0 & 15 & 55 & 51.984 & 26.207 \\ 
 & Oracle       & 42.3\% & 0.000 & 100.0 & 43.7 & 13.3 & 30.3 & 18.4 & 0 & 2562 & 0 & 0 & 64 & 64 & 28.812 & 14.099 \\ \hline
\multirow{3}{*}{\STAB{\rotatebox[origin=c]{90}{Tst1}}} 
 & Naive        &  72.1\% & 8.828 & 92.8 & 78.2 & 78.2 & 92.8 & 84.9 & 55 & 196 & 0 & 0 & 0 & 35 & 19.556 & 9.778 \\
 & FRCNN-H      &  37.8\% & 6.992 & 100.0 & 37.8 & 37.8 & 100.0 & 54.8 & 0 & 560 & 0 & 0 & 0 & 7 & 42.500 & 20.000 \\ 
 & Oracle       & 100.0\% & 0.000 & 100.0 & 100.0 & 100.0 & 100.0 & 100.0 & 0 & 0 & 1 & 0 & 0 & 0 & 900.000 & 900.000 \\ \hline
\multirow{3}{*}{\STAB{\rotatebox[origin=c]{90}{Tst2}}} 
 & Naive        & 75.7\% & 6.560 & 95.6 & 79.4 & 79.4 & 95.6 & 86.7 & 66 & 371 & 1 & 0 & 0 & 38 & 35.725 & 18.089 \\
 & FRCNN-H      &  7.8\% & 6.187 & 100.0 & 7.9 & 4.8 & 61.3 & 9.0 & 0 & 1658 & 0 & 2 & 2 & 10 & 11.833 & 5.462 \\
 & Oracle       & 82.2\% & 0.000 & 100.0 & 83.2 & 24.3 & 29.2 & 26.6 & 0 & 302 & 2 & 0 & 18 & 18 & 74.900 & 37.450 \\ \hline
\multirow{3}{*}{\STAB{\rotatebox[origin=c]{90}{Tst5}}} 
 & Naive        & 39.7\% & 9.665 & 77.9 & 56.1 & 38.6 & 53.6 & 44.9 & 717 & 1977 & 0 & 0 & 21 & 229 & 10.736 & 5.403 \\
 & FRCNN-H      & 50.1\% & 5.460 & 99.4 & 50.9 & 27.2 & 53.2 & 36.0 & 13 & 2209 & 0 & 0 & 23 & 91 & 23.619 & 11.749 \\
 & Oracle       & 65.7\% & 0.000 & 100.0 & 66.8 & 26.8 & 40.1 & 32.2 & 0 & 1492 & 1 & 0 & 50 & 50 & 54.691 & 27.345 \\ \hline
\multirow{3}{*}{\STAB{\rotatebox[origin=c]{90}{Tst10}}} 
 & Naive        & 42.1\% & 10.218 & 78.3 & 59.0 & 45.4 & 60.3 & 51.8 & 1467 & 3693 & 0 & 0 & 47 & 339 & 15.034 & 7.506 \\
 & FRCNN-H      & 39.8\% & 5.488 & 99.9 & 40.3 & 19.6 & 48.8 & 28.0 & 3 & 5377 & 0 & 0 & 37 & 120 & 27.656 & 13.620 \\
 & Oracle       & 64.7\% & 0.000 & 100.0 & 66.0 & 20.0 & 30.3 & 24.1 & 0 & 3059 & 1 & 0 & 119 & 119 & 46.054 & 22.589 \\ \hline
\end{tabular}%
}
\caption{Full metric evaluation of the 2D tracking in the front-view, after 3D tracklet association, on all sequences.}
\label{tab:trackMeasuresFront3D}
\end{table*}

\begin{table*}[t!]
\centering
\resizebox{\linewidth}{!}{%
\begin{tabular}{clccccccccccccccc}
\hline
 & Method & MOTA $\uparrow$ & MOTP $\downarrow$ & Prc. $\uparrow$ & Rcll. $\uparrow$ & ID-Rcll. $\uparrow$ & ID-Prc. $\uparrow$ & ID-F1 $\uparrow$ & FP $\downarrow$ & FN $\downarrow$ & MT $\uparrow$ & ML $\downarrow$ & ID Sw. $\downarrow$ & Frag. $\downarrow$ & $\textrm{MTBF}_{s}$ $\uparrow$ & $\textrm{MTBF}_{m}$ $\uparrow$ \\ \hline
\multirow{3}{*}{\STAB{\rotatebox[origin=c]{90}{Trn2}}} 
 & Naive        & 41.1\% & 0.198 & 85.0 & 50.5 & 9.1 & 15.3 & 11.4 & 1285 & 7118 & 0 & 0 & 60 & 586 & 12.344 & 6.172 \\
 & FRCNN-H      & 73.7\% & 0.066 & 96.9 & 76.3 & 18.8 & 23.8 & 21.0 & 352 & 3407 & 0 & 0 & 29 & 215 & 50.548 & 25.333 \\
 & Oracle       & 44.8\% & 0.000 & 100.0 & 46.2 & 2.4 & 5.3 & 3.4 & 0 & 7738 & 0 & 0 & 202 & 202 & 32.539 & 16.190 \\ \hline
\multirow{3}{*}{\STAB{\rotatebox[origin=c]{90}{Trn5}}} 
 & Naive        & 40.3\% & 0.169 & 83.5 & 50.5 & 25.5 & 42.1 & 31.8 & 448 & 2218 & 0 & 0 & 12 & 134 & 16.309 & 8.039 \\
 & FRCNN-H      & 63.5\% & 0.064 & 92.1 & 69.7 & 37.1 & 49.1 & 42.3 & 269 & 1358 & 1 & 0 & 9 & 67 & 43.431 & 21.867 \\
 & Oracle       & 36.7\% & 0.000 & 100.0 & 37.9 & 11.0 & 28.9 & 15.9 & 0 & 2784 & 0 & 0 & 53 & 53 & 29.328 & 14.175 \\ \hline
\multirow{3}{*}{\STAB{\rotatebox[origin=c]{90}{Val2}}} 
 & Naive        & 59.7\% & 0.163 & 92.2 & 65.4 & 48.5 & 68.3 & 56.7 & 200 & 1244 & 0 & 0 & 5 & 86 & 26.682 & 13.341 \\
 & FRCNN-H      & 71.4\% & 0.062 & 96.3 & 74.4 & 61.4 & 79.5 & 69.3 & 103 & 919 & 0 & 0 & 4 & 36 & 70.342 & 36.122 \\
 & Oracle       & 74.1\% & 0.000 & 100.0 & 75.1 & 12.8 & 17.0 & 14.6 & 0 & 894 & 0 & 0 & 36 & 36 & 71.000 & 36.459 \\ \hline
\multirow{3}{*}{\STAB{\rotatebox[origin=c]{90}{Val5}}} 
 & Naive        & 28.4\% & 0.204 & 78.8 & 39.1 & 23.2 & 46.8 & 31.1 & 477 & 2764 & 0 & 0 & 8 & 93 & 18.122 & 9.015 \\
 & FRCNN-H      & 61.3\% & 0.070 & 95.4 & 64.6 & 33.7 & 49.7 & 40.2 & 143 & 1607 & 1 & 0 & 9 & 60 & 45.123 & 22.562 \\
 & Oracle       & 42.4\% & 0.000 & 100.0 & 43.8 & 13.3 & 30.3 & 18.5 & 0 & 2552 & 0 & 0 & 64 & 64 & 28.812 & 14.618 \\ \hline
\multirow{3}{*}{\STAB{\rotatebox[origin=c]{90}{Tst1}}} 
 & Naive        &  77.6\% & 0.152 & 96.0 & 80.9 & 80.9 & 96.0 & 87.8 & 30 & 171 & 1 & 0 & 0 & 28 & 25.000 & 12.500 \\
 & FRCNN-H      &  30.2\% & 0.105 & 100.0 & 30.2 & 30.2 & 100.0 & 46.4 & 0 & 625 & 0 & 0 & 0 & 15 & 16.938 & 8.212 \\
 & Oracle       & 100.0\% & 0.000 & 100.0 & 100.0 & 100.0 & 100.0 & 100.0 & 0 & 0 & 1 & 0 & 0 & 0 & 900.000 & 900.000 \\ \hline
\multirow{3}{*}{\STAB{\rotatebox[origin=c]{90}{Tst2}}} 
 & Naive        & 77.6\% & 0.138 & 96.9 & 80.2 & 80.2 & 96.9 & 87.7 & 46 & 353 & 1 & 0 & 0 & 44 & 31.022 & 15.856 \\
 & FRCNN-H      &  5.7\% & 0.133 & 100.0 & 5.8 & 3.5 & 60.2 & 6.6 & 0 & 1677 & 0 & 2 & 2 & 17 & 5.421 & 2.641 \\
 & Oracle       & 80.6\% & 0.000 & 100.0 & 81.6 & 23.9 & 29.3 & 26.3 & 0 & 328 & 2 & 0 & 18 & 25 & 53.778 & 27.396 \\ \hline
\multirow{3}{*}{\STAB{\rotatebox[origin=c]{90}{Tst5}}} 
 & Naive        & 39.7\% & 0.168 & 80.0 & 53.1 & 38.0 & 57.3 & 45.7 & 589 & 2078 & 0 & 0 & 9 & 186 & 12.340 & 6.219 \\
 & FRCNN-H      & 40.0\% & 0.099 & 98.0 & 41.3 & 23.2 & 55.1 & 32.6 & 37 & 2605 & 0 & 0 & 17 & 117 & 15.000 & 7.469 \\
 & Oracle       & 66.7\% & 0.000 & 100.0 & 67.8 & 27.2 & 40.1 & 32.4 & 0 & 1427 & 1 & 0 & 50 & 50 & 54.691 & 28.112 \\ \hline
\multirow{3}{*}{\STAB{\rotatebox[origin=c]{90}{Tst10}}} 
 & Naive        & 48.2\% & 0.153 & 86.7 & 57.1 & 46.3 & 70.3 & 55.8 & 780 & 3824 & 0 & 0 & 16 & 273 & 18.007 & 8.925 \\
 & FRCNN-H      & 25.2\% & 0.099 & 97.0 & 26.3 & 14.1 & 52.0 & 22.2 & 72 & 6571 & 0 & 3 & 32 & 225 & 9.996 & 4.904 \\
 & Oracle       & 65.2\% & 0.000 & 100.0 & 66.6 & 20.2 & 30.3 & 24.3 & 0 & 2982 & 1 & 0 & 119 & 119 & 46.031 & 23.105 \\ \hline
\end{tabular}%
}
\caption{Full metric evaluation of the 3D tracklets generated from the 3D tracklet association, on all sequences.}
\label{tab:trackMeasures3DTracklets}
\end{table*}

\begin{table*}[t!]
\centering
\resizebox{\linewidth}{!}{%
\begin{tabular}{clccccccccccccccc}
\hline
 & Method & MOTA $\uparrow$ & MOTP $\downarrow$ & Prc. $\uparrow$ & Rcll. $\uparrow$ & ID-Rcll. $\uparrow$ & ID-Prc. $\uparrow$ & ID-F1 $\uparrow$ & FP $\downarrow$ & FN $\downarrow$ & MT $\uparrow$ & ML $\downarrow$ & ID Sw. $\downarrow$ & Frag. $\downarrow$ & $\textrm{MTBF}_{s}$ $\uparrow$ & $\textrm{MTBF}_{m}$ $\uparrow$ \\ \hline
\multirow{3}{*}{\STAB{\rotatebox[origin=c]{90}{Trn2}}} 
 & Naive        & 41.4\% & 0.198 & 85.3 & 50.4 & 28.5 & 48.3 & 35.9 & 1250 & 7133 & 0 & 0 & 40 & 573 & 12.597 & 6.298 \\
 & FRCNN-H      & 73.8\% & 0.066 & 96.9 & 76.3 & 44.0 & 55.9 & 49.2 & 352 & 3407 & 0 & 0 & 14 & 216 & 50.317 & 25.216 \\
 & Oracle       & 46.2\% & 0.000 & 100.0 & 46.2 & 46.2 & 100.0 & 63.2 & 0 & 7738 & 0 & 0 & 0 & 202 & 32.539 & 16.190 \\ \hline
\multirow{3}{*}{\STAB{\rotatebox[origin=c]{90}{Trn5}}} 
 & Naive        & 40.4\% & 0.170 & 83.8 & 50.3 & 32.5 & 54.2 & 40.7 & 436 & 2228 & 0 & 0 & 7 & 135 & 16.121 & 7.947 \\
 & FRCNN-H      & 63.5\% & 0.064 & 92.1 & 69.7 & 40.9 & 54.1 & 46.6 & 269 & 1359 & 1 & 0 & 7 & 66 & 44.028 & 22.170 \\
 & Oracle       & 37.9\% & 0.000 & 100.0 & 37.9 & 37.9 & 100.0 & 55.0 & 0 & 2784 & 0 & 0 & 0 & 53 & 29.328 & 14.175 \\ \hline
\multirow{3}{*}{\STAB{\rotatebox[origin=c]{90}{Val2}}} 
 & Naive        & 59.6\% & 0.163 & 92.2 & 65.3 & 50.8 & 71.7 & 59.4 & 199 & 1248 & 0 & 0 & 3 & 82 & 27.905 & 13.870 \\
 & FRCNN-H      & 71.5\% & 0.062 & 96.3 & 74.4 & 71.7 & 92.8 & 80.9 & 103 & 920 & 0 & 0 & 2 & 35 & 72.216 & 37.111 \\
 & Oracle       & 75.1\% & 0.000 & 100.0 & 75.1 & 75.1 & 100.0 & 85.8 & 0 & 894 & 0 & 0 & 0 & 36 & 71.000 & 36.459 \\\hline
\multirow{3}{*}{\STAB{\rotatebox[origin=c]{90}{Val5}}} 
 & Naive        & 28.5\% & 0.204 & 78.9 & 39.1 & 23.2 & 46.8 & 31.1 & 476 & 2764 & 0 & 0 & 7 & 93 & 18.122 & 9.015 \\
 & FRCNN-H      & 61.3\% & 0.070 & 95.4 & 64.6 & 40.2 & 59.4 & 48.0 & 143 & 1607 & 1 & 0 & 5 & 60 & 45.123 & 22.562 \\
 & Oracle       & 43.8\% & 0.000 & 100.0 & 43.8 & 43.8 & 100.0 & 60.9 & 0 & 2552 & 0 & 0 & 0 & 64 & 28.812 & 14.618 \\ \hline
\multirow{3}{*}{\STAB{\rotatebox[origin=c]{90}{Tst1}}} 
 & Naive        &  77.6\% & 0.152 & 96.0 & 80.9 & 80.9 & 96.0 & 87.8 & 30 & 171 & 1 & 0 & 0 & 28 & 25.000 & 12.500 \\
 & FRCNN-H      &  30.2\% & 0.105 & 100.0 & 30.2 & 30.2 & 100.0 & 46.4 & 0 & 625 & 0 & 0 & 0 & 15 & 16.938 & 8.212 \\
 & Oracle       & 100.0\% & 0.000 & 100.0 & 100.0 & 100.0 & 100.0 & 100.0 & 0 & 0 & 1 & 0 & 0 & 0 & 900.000 & 900.000 \\ \hline
\multirow{3}{*}{\STAB{\rotatebox[origin=c]{90}{Tst2}}} 
 & Naive        & 77.6\% & 0.138 & 96.9 & 80.2 & 80.2 & 96.9 & 87.7 & 46 & 353 & 1 & 0 & 0 & 44 & 31.022 & 15.856 \\
 & FRCNN-H*     &  5.7\% & 0.133 & 100.0 & 5.8 & 3.5 & 60.2 & 6.6 & 0 & 1677 & 0 & 2 & 2 & 17 & 5.421 & 2.641 \\
 & Oracle       & 81.6\% & 0.000 & 100.0 & 81.6 & 81.6 & 100.0 & 89.9 & 0 & 328 & 2 & 0 & 0 & 25 & 53.778 & 27.396 \\ \hline
\multirow{3}{*}{\STAB{\rotatebox[origin=c]{90}{Tst5}}}
 & Naive        & 39.7\% & 0.168 & 80.0 & 53.1 & 43.4 & 65.4 & 52.2 & 588 & 2079 & 0 & 0 & 7 & 185 & 12.400 & 6.249 \\
 & FRCNN-H      & 40.2\% & 0.099 & 98.0 & 41.2 & 29.8 & 70.9 & 41.9 & 37 & 2609 & 0 & 0 & 7 & 115 & 15.217 & 7.577 \\ 
 & Oracle       & 67.8\% & 0.000 & 100.0 & 67.8 & 67.8 & 100.0 & 80.8 & 0 & 1427 & 1 & 0 & 0 & 50 & 54.691 & 28.112 \\ \hline
\multirow{3}{*}{\STAB{\rotatebox[origin=c]{90}{Tst10}}} 
 & Naive        & 48.3\% & 0.153 & 86.9 & 57.1 & 48.5 & 73.8 & 58.5 & 768 & 3829 & 0 & 0 & 11 & 268 & 18.313 & 9.075 \\
 & FRCNN-H*     & 25.2\% & 0.099 & 97.0 & 26.3 & 14.1 & 52.0 & 22.2 & 72 & 6571 & 0 & 3 & 32 & 225 & 9.996 & 4.904 \\
 & Oracle       & 66.6\% & 0.000 & 100.0 & 66.6 & 66.6 & 100.0 & 79.9 & 0 & 2982 & 1 & 0 & 0 & 119 & 46.031 & 23.105 \\ \hline
\end{tabular}%
}
\caption{Full metric evaluation of the final 3D tracks generated from the 3D track association, on all sequences. An * indicates that the method did not complete the 3D Tracklet Association step for the sequence, as there were not enough concurrent 3D tracklets at any one point in the sequence. In this case, the 3D tracklet results are reported.}
\label{tab:trackMeasures3DTracks}
\end{table*}

\end{document}